\documentclass[10pt]{article}

\usepackage[preprint]{tmlr}
\usepackage{amsmath,amssymb}
\usepackage{booktabs}
\usepackage{array}
\usepackage{tabularx}
\usepackage{longtable}
\usepackage{enumitem}
\usepackage{graphicx}
\usepackage[dvipsnames]{xcolor}
\usepackage{multirow}
\usepackage{caption}
\usepackage{subcaption}
\usepackage[section]{placeins}
\usepackage{soul}
\usepackage{hyperref}
\usepackage{microtype}
\hypersetup{colorlinks=true,linkcolor=NavyBlue,citecolor=NavyBlue,urlcolor=NavyBlue}


\title{Why Units Die in the Layer After a Frozen Random Encoder, and What the Number of Live Units Measures}

\author{\name Scott M. Norton \email nortonsm@urgrad.rochester.edu \\
        \addr Independent Researcher}

\def\month{11}
\def\year{2026}

\newcommand{\xlabel}[2]{#1}
\begin{document}

\maketitle

\begin{abstract}
Reinforcement-learning agents can learn Atari games through a convolutional encoder whose random weights are never trained. They reach expert play in Pong and, without random action repeats, in Breakout, but training takes far longer and few ReLU units in the layer after the encoder stay alive. A dead unit outputs zero for every input and cannot learn. In earlier work we took the number of live units to measure the game's complexity. Here we ask why units die and correct that conclusion. We find that more units die as the layer's inputs lie farther from zero relative to how much they vary. We call this ratio the offset. Earlier work showed this in sparse autoencoders at the start of training. We saw it after training in agents, image classifiers, a reservoir computer and sparse autoencoders on GPT-2. A random encoder leaves the offset of its input roughly unchanged, and a formula predicted its median offset within 25 percent in all 44 cases. So a game with a bright, mostly unchanging screen gives the next layer a large offset. In Pong, subtracting the average screen raised the number of live units from between 3 and 6 of 64 to between 59 and 63, so the number alone cannot show a game's complexity. On unaltered screens none of the agents reached a training score of 18 in 50 million steps. With the subtraction all did within 5 million, about as fast as a trained encoder on unaltered screens. In Breakout the subtraction left more units alive, but average training scores did not clearly improve, and a large gap to a trained encoder remained. So we do not treat their number as something to raise for its own sake. The evidence is from small networks and three games, and in agents and image classifiers changing the offset also changed the size of the inputs. We suggest measuring the offset before training a layer on a fixed front end and, if it is large, trying to subtract the mean input.
\end{abstract}

\section{Introduction}

In many machine-learning systems the first part of the network is fixed, and only the later part is trained. A reservoir computer sends its input through a recurrent network whose weights are chosen at random and never changed, and it trains only the layer that comes after \citep{jaeger2001echo}. Random-feature models work in the same way \citep{rahimi2007random,huang2006extreme}. In robot learning it is common to take an image encoder that was trained on other data, freeze it, and train a small network on its output \citep{parisi2022unsurprising,nair2022r3m,majumdar2023vc1}. Sparse autoencoders are trained to rewrite the internal activations of a language model as a sum of a small number of learned items, and the language model itself stays fixed \citep{bricken2023monosemanticity,gao2025scaling}. Some recent studies also use an encoder left at its random starting weights as a control for a pretrained one \citep{berger2026ecg,isiacik2026timeseries}.

In all of these systems, the first layer that is trained has no control over its inputs. It has to work with whatever the fixed part of the network produces. We call that fixed part the front end, and this paper asks what a fixed front end does to the layer that is trained on top of it. Our answer, for the systems we tested, is that the inputs to that layer can lie far from zero compared with how much they vary, and that when they do, many of the layer's units stop working.

To say what ``stop working'' means, we need one fact about the units themselves. Most networks, including ours, are built from rectified linear units, or ReLUs. A ReLU multiplies each of its inputs by a weight, adds the results together and adds a constant called the bias. We call this total the unit's input sum. If the input sum is positive, the unit outputs it. If the input sum is negative, the unit outputs zero. Now suppose the input sum is negative for every example the network ever sees. The unit then outputs zero every time. A unit that outputs zero also receives no training signal, so its weights stop changing and ordinary training cannot switch it back on. Such a unit is called dead, and a unit that is not dead is called alive. The number of units in a layer that are still alive at the end of training is the number of live units of our title. In the text we also call it simply the count.

Dead units matter for two reasons. The first is that a dead unit is part of the network that is no longer being used, and in sparse autoencoders this is a well-known practical problem \citep{bricken2023monosemanticity,gao2025scaling}. The second is that people use the number of dead units as a measurement. In reinforcement learning, a growing number of dead units, or of dormant units, is widely taken as a sign that a network is losing its ability to keep learning \citep{sokar2023dormant,lyle2023plasticity,dohare2024plasticity}. (A dormant unit is one whose average output is very small compared with the other units in its layer.) In an earlier study of our own \citep{norton2026frozen} we interpreted the number in yet another way. We found that many units died in agents with a random front end, more in some games than in others, and we concluded that the number left alive measures how complex the game is. One aim of this paper is to check that conclusion. We find that the evidence we gave for it, the order of the counts across games, does not support it (Section 4.5).

Our main test case is the same kind of agent. These agents learn to play Atari video games, a standard benchmark in reinforcement learning, by trial and error from the score. At every step the agent is shown the screen of the game as a stack of four frames. The first three layers of the network are convolutional, and together we call them the encoder. The encoder turns each screen into a list of 3,136 numbers, and the layer after it receives those numbers as its inputs. In an ordinary agent the encoder is trained along with everything else. In ours its weights are set at random at the start and never changed, and only the layers after it are trained. We call this a frozen random encoder (Section 3.1).

Such an agent can still learn to play well. In the earlier study, all five Pong agents with a frozen random encoder finished with scores between +16 and +20, in a game where the score runs from $-21$ to $+21$. In Breakout, three agents scored 221, 303 and 351, compared with 388 for an agent whose encoder was trained. The Breakout scores were obtained with sticky actions turned off. With sticky actions, the game sometimes repeats the agent's previous action in place of the new one. The drawback of a frozen encoder is training time. The Pong agents needed at least 8.9 times as many agent steps as an agent with a trained encoder to reach an average score of zero, where an agent step is one action chosen by the agent. The first trained layer of these agents has 64 units. Few of them stayed alive. In one Pong agent that finished at +18.8, only three were still alive. We state this plainly, because dead units are usually taken as a sign of a network in trouble. A network that ends with few live units is not necessarily a poor one. The Pong agents reached expert play with between 3 and 8 live units of the 64, and the Breakout agents, with sticky actions off, with between 20 and 23. As we have said, these agents also took far longer to train.

A frozen random encoder is a convenient system for studying dead units, for two reasons. First, the inputs to the first trained layer are fixed before training begins, so we can measure them, change them on purpose and see what happens. Second, the encoder's weights are random and contain no information about the game, so we can work out mathematically what the encoder does to its input.

\begin{figure}[htbp]
\centering
\includegraphics[width=\textwidth,height=0.80\textheight,keepaspectratio]{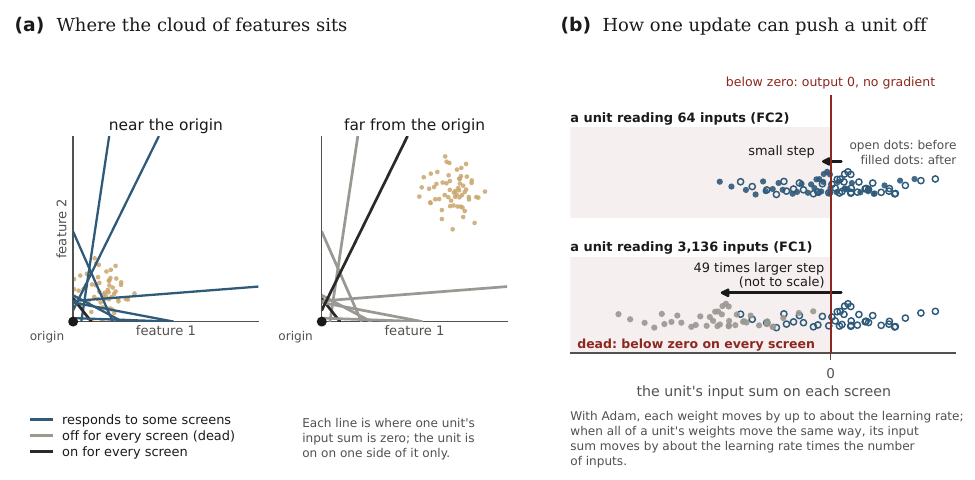}
{\captionsetup{font=small}\caption{How a unit of the first dense layer dies, drawn as a schematic and not from data. (a) Dots are screens, and each line marks where one unit's input is zero. On the left the screens form a cloud near the origin, and most lines cut through it. On the right the cloud is far from the origin compared with its spread, most lines miss it, and a unit that is off for every screen is dead. (b) One update moves a unit's input on every screen at once, and a unit with more inputs can move further.}\label{fig:schematic}}
\end{figure}

We find that the number of units that stay alive depends on two quantities, and both can be measured before training starts (Figure \ref{fig:schematic}). The evidence is stronger for the first than for the second. The first is the \textbf{offset}. We describe it with a picture. Think of the inputs that the layer receives for one example as a point, and of the inputs for all the examples as a collection of points. In our Atari agents an example is one screen of the game, that is, one stack of four frames. The encoder turns the screen into 3,136 numbers, and these numbers are the inputs of the first trained layer. Taken together they mark one point, in a space that has 3,136 directions. Each screen that the agent meets gives one such point, and the screens of a whole game give a collection of points. The offset is the distance from zero to the centre of that collection, divided by the spread of the collection, which is the square root of its total variance (Section 3.2). When the offset is small, the points are scattered around zero. When it is large, the points form a tight group far from zero. Pong is a case of the second kind. Most of a Pong screen is a bright background that stays the same from one moment to the next, so the screens themselves have a large offset (Section 5.1). We show in Section 5.2 that the output of a random encoder has roughly the same offset as its input. The points of the Pong screens therefore form a tight group far from zero. To give a sense of scale, the GPT-2 activations that we use in Section 4.3 have an offset of 0.904, and the outputs of random encoders on Pong screens have a median offset of 27.54.

The offset matters because of how a ReLU divides up its inputs. Each unit is on for the points on one side of a boundary and off for the points on the other side. In our agents this means that the unit is on for some screens and off for the others. At the start of training the boundary is placed at random. If the points are scattered around zero, a random boundary usually has some points on each side, so the unit is on for some examples and off for others. If the points form a tight group far from zero, a random boundary usually misses the group altogether. The unit is then on for every example or off for every example, and a unit that is off for every example receives no training signal and stays off. In our agents this means a unit that is off for every screen. Simon, Adams and Zou \citeyearpar{simon2026outliers} worked this out for sparse autoencoders, and gave a formula for the fraction of units that are dead at the start of training.

The second quantity is the \textbf{step size}, by which we mean how far one training step can move a unit's input sum. Our networks are trained with the Adam optimiser \citep{kingma2015adam}. Under Adam, one step changes each weight by at most about the learning rate. If the changes to all of a unit's weights push its input sum in the same direction, they add up, and the input sum moves by about the learning rate multiplied by the number of inputs the unit has. That number is called the unit's \textbf{fan-in}. A unit with many inputs can therefore be moved a long way by a single step. Yang et al. \citeyearpar{yang2021tensor} use the same scaling to argue that the learning rate under Adam should be reduced in proportion to the fan-in, and Kosson et al. \citeyearpar{kosson2026weightdecay} measured that the weight changes do line up with the inputs early in training.

Both quantities build on earlier work. The offset was worked out by Simon, Adams and Zou for sparse autoencoders at the start of training. The step size is our own derivation from two established results, the limit on how far Adam moves each weight and the observation that early weight changes line up with the inputs (Section 4.4). But as far as we could find, earlier work leaves three questions open.

The first is what happens after the start of training. Simon, Adams and Zou showed that a large offset leaves units dead at the start, and they argue that those units recover only slowly. They report that units can also die later, during training, but they do not relate those later deaths to the offset, and their analysis is of sparse autoencoders. One other analysis, by Shvetsov et al. \citeyearpar{shvetsov2026bugfeature}, links activations that have a positive mean to a drift of the weights during training (Section 2). The second question is what causes units to die in the layer after an encoder. Sokar and Castro \citeyearpar{sokar2025gap} found that in widened agents a large share of the dormant units are in that layer, but they did not test the cause. The third question is what a frozen random encoder gives to the layer after it. Such encoders have been found to do much worse than trained ones in Atari games \citep{stooke2021decoupling}, and in our earlier study they needed far longer training to reach expert play. We did not find earlier work that relates the offset of their output to the offset of their input, changes that offset before training, and looks at what happens to the next layer.

The paper reports four findings.

\begin{enumerate}[noitemsep,topsep=2pt]
\item The number of live units depends on the offset and, on weaker evidence, on the step size, and on its own it does not tell us how complex the task is. When we raised the offset before training, fewer units stayed alive. We saw this in agents playing Breakout and Space Invaders, in image classifiers, in a reservoir computer and in sparse autoencoders on GPT-2. The clearest case is Pong, where we lowered the offset. There we subtracted the average screen from every screen, a step we call centring, which lowers the offset without changing the game or the encoder. The number of live units went from between 3 and 6 of the 64 to between 59 and 63, by the 1 percent rule, which is our main counting rule (Section 3.3). Making a task harder did change the number of live units in image classifiers, but not in a consistent direction (Section 4).
\item The output of a random encoder has roughly the same offset as its input. We give a formula that predicts the offset of the output from a sample of the inputs and the sizes of the layers, without drawing any weights. It predicted the median offset over random encoders to within 25 percent in all 44 conditions we tried, which included game screens and CIFAR-10 photographs. The test was fixed in advance for 33 of the conditions, and the other 11 were added afterwards with the same 25 percent band (Section 5).
\item Once the average screen was subtracted, agents with a frozen encoder learned Pong about as fast as agents with a trained encoder did on unaltered screens, that is, on the screens exactly as the game produces them. We measure speed by the agent step at which the score during training first reached 18, and ``about as fast'' means that the two differed by no more than repeated runs of the same agent differ. In single runs, a trained encoder on centred screens was faster than either. In Breakout the frozen encoder stayed far behind. We also found that a trained encoder lowered its own offset early in training (Section 6).
\item Having more live units did not always mean better learning. This finding is about agents with a frozen encoder. When we divided the learning rate of the first trained layer by 49, between 36 and 46 of its units stayed alive in Breakout, compared with between 9 and 26 at the usual learning rate (1 percent rule), and yet those agents scored lower. A smaller learning rate may itself slow learning, so this test is not a clean one (Section 7). The reverse is also true in our data. Having few live units did not rule out playing well, because the agents of the earlier study reached expert play with few live units, after much longer training (Section 6.1).
\end{enumerate}

These findings address the three questions, within the limits given in Sections 4 to 7. On the first, a higher offset left fewer units alive at the end of training in agents, image classifiers and a reservoir computer, so the offset matters outside sparse autoencoders. In sparse autoencoders at four and eight times the unaltered offset, and at a learning rate of 0.001, most of the units that were dead at the end had been alive at the start, so those units died during training, most of them within the first 50 steps. In the agents we cannot say how much of the loss comes before training starts and how much during training (Section 8.3). On the second, we propose two quantities that the number of live units depends on in the layer after a frozen encoder, and we changed each of them on purpose. The evidence for the offset points the same way in each kind of system we tried. The evidence for the step size is weaker (Section 4.4). On the third, we show that a random encoder gives the next layer roughly the offset of its own input, and we give a formula that predicts it. We also report what happened to learning when the average screen was subtracted. In our Pong runs, agents with a frozen encoder did not reach a training score of 18 in 50 million agent steps on unaltered screens. With the average screen subtracted they reached it within 5 million, which is about as fast as agents with a trained encoder on unaltered screens. In Breakout the subtraction did not close the gap.

Finally, we correct a conclusion of our own. The earlier study took the number of live units as a measure of how complex a game is. But the same game, with the same encoders, ended with between 3 and 6 live units or with between 59 and 63, depending on the offset. So the order of the counts across games in that study is not evidence that the games differ in complexity in that order. One of our tests in image classifiers does fit the earlier view, and we report it in Section 4.5. Table \ref{tab:adds} in Section 2 sets what we found beside what was known and what was missing.

In practice we suggest three things. Measure the offset of a layer's inputs before training, which takes one pass over a sample of them. If the offset is large, try subtracting the mean input first. And report the offset with any count of dead or dormant units, because two networks can have different numbers of dead units without differing in how well they learned (Section 8.1).

The evidence has limits, and Section 8.3 collects them. The main experiments use small networks and three Atari games. Most comparisons between agents use a single run for each encoder in each setting. Where we did repeat runs, three runs of the same Breakout encoder differed by as many as 7 live units. The tests outside games are small. In the agents and the image classifiers, our changes to the offset also changed the overall size of the layer's inputs, and we did not separate the two. Several of the tests whose pass and fail levels we had written down in advance failed or passed only in part. One of them checks whether the units we count as alive are really the ones an agent uses to play. It failed in three experiments and was not run in several others (Section 3.3; Supplement S5 and S18). The Supplement is a separate document provided with this paper.

\section{Background}

This section reviews the earlier work that the paper builds on, in four groups. For each group we say what was known and what we could not find. Table \ref{tab:adds} at the end of the section sets what was known beside what was missing and what this paper found.

The first group is about dead and dormant units. Sokar et al. \citeyearpar{sokar2023dormant} showed that dormant units build up in reinforcement-learning agents as training goes on, and that resetting those units helps an agent to keep learning. Lyle et al. \citeyearpar{lyle2023plasticity} studied how networks lose plasticity, which is the ability to keep learning from new data. They showed that the number of dead units is not a reliable guide to how much plasticity a network has lost. In later work they argue that a loss of plasticity can have several independent causes \citep{lyle2024disentangling}. Sokar and Castro \citeyearpar{sokar2025gap} looked at where in the network the dormant units are. They studied agents whose networks had been made four times wider, and found that a large share of the dormant units were in one layer, the dense layer that takes the flattened output of the encoder as its input. They also found fewer dormant units when they replaced the flattening step with global average pooling, which gives that layer far fewer inputs. Finally, Liu et al. \citeyearpar{liu2025grama} argue that a measure of dormancy based on a unit's activations loses statistical power in more complex network architectures, and they propose a measure based on the size of the gradients. What this work leaves open is the cause. Sokar and Castro located dormant units in the layer after the encoder, but they did not test why the units in that layer stop working. Trumpp et al. \citeyearpar{trumpp2026resolution} suggest a reason, that the weights of that dense layer grow fast with the size of the image, but they do not test it.

The second group is about inputs that are far from zero. The advice to centre the inputs of every layer, that is, to subtract their mean, is decades old \citep{lecun1998efficient}. One reason it matters is that the output of a ReLU is never negative. The average output of a layer of ReLUs is therefore positive, and this positive average acts like an extra bias on the next layer. Clevert et al. \citeyearpar{clevert2016elu} call this a bias shift. Brock et al. \citeyearpar{brock2021characterizing}, who studied networks without normalising layers, call it a mean shift. The work closest to ours is by Simon, Adams and Zou \citeyearpar{simon2026outliers}, who studied sparse autoencoders. They showed that the quantity we call the offset predicts the fraction of units that are dead at the start of training, and that subtracting the mean of the inputs largely removes those dead units. Their analysis is of the start of training. They also argue that units which are dead at the start recover only slowly when the offset is large. They report that a strong penalty or a high learning rate can make units die during training as well, and that a high learning rate can do this even when the inputs are centred. They do not relate those later deaths to the offset. Shvetsov et al. \citeyearpar{shvetsov2026bugfeature} show that when activations have a positive mean, the gradients of the weights that follow all have one sign at the start of training, and that the resulting drift of the weights grows with the learning rate. Apart from that analysis, we did not find work that ties units dying during training to the offset of a layer's inputs. We also did not find an experiment that raises and lowers the offset in a reinforcement-learning agent. These are the two gaps we address. Lyle et al. \citeyearpar{lyle2024normalization} found that centring does little against dead units, and that the benefit of layer normalisation against saturation comes from rescaling. We compare these operations after a fixed front end in Section 4.3.

The third group is about the size of training steps. Under Adam, one step changes each weight by at most about the learning rate \citep{kingma2015adam}. Yang et al. \citeyearpar{yang2021tensor} derive a rule in which the learning rate under Adam is reduced in proportion to the fan-in, and Kosson et al. \citeyearpar{kosson2026weightdecay} measured that weight changes line up with the inputs early in training. It is also known that the learning rate affects how many units die \citep{dufortlabbe2025demp}, and Sokar and Castro \citeyearpar{sokar2025gap} found that the layer with the most inputs has the most dormant units and that pooling helps. What we did not find is work that relates the number of live units to the learning rate multiplied by the fan-in. The rule itself is not ours. We apply it to a new outcome in Section 4.4.

The fourth group is about front ends that are fixed or random. Random filters, random projections and reservoirs have been studied for a long time \citep{jarrett2009best,saxe2011random,rahimi2007random,huang2006extreme,jaeger2001echo,maass2002real}. In reinforcement learning, frozen random networks have been used in methods for exploration \citep{burda2019exploration,burda2019largescale,seo2021state}, as the front end of a controller trained by an evolution strategy \citep{chang2020reinforcement}, and as baselines in robot learning \citep{parisi2022unsurprising,majumdar2023vc1}. Stooke et al. \citeyearpar{stooke2021decoupling} found that frozen random encoders did much worse than trained ones in four Atari games. Our earlier study \citep{norton2026frozen} found that agents with a frozen random encoder can nevertheless reach expert play in Pong and, without sticky actions, in Breakout. The outputs of networks with random weights have also been analysed mathematically \citep{cho2009kernel,garrigaalonso2019deep,novak2019bayesian}, and Section 5.3 builds on that work. What we did not find is work that relates the offset of a random encoder's output to the offset of its input, changes that offset before training, and counts how many units of the next layer are alive at the end. Our search cannot rule out that such work exists. That is how we use a frozen random encoder in this paper.

It is also useful to know how widely used training code treats its inputs, because this affects the offset that the first layers receive. The Atari implementations of PPO, the training method we use (Section 3.1), in CleanRL \citep{huang2022cleanrl} and Stable-Baselines3 \citep{raffin2021stable} divide the pixel values by 255. They do not subtract a mean, and they have no normalising layer. The published code of DrQ-v2 \citep{yarats2022drqv2} and of DreamerV3 \citep{hafner2025dreamerv3} goes further. Both also subtract 0.5. DrQ-v2 then places a normalising layer before the first nonlinearity after the encoder, and DreamerV3 places one before every nonlinearity. SimBa \citep{lee2025simba} normalises its observations using running statistics, and its authors report that performance drops when this step is removed. Andrychowicz et al. \citeyearpar{andrychowicz2021matters} carried out a large study of on-policy agents on continuous-control tasks. They found that normalising the observations was among the choices that mattered most, and they recommend always doing it. Subtracting a background or a running mean from the observations is itself not new \citep{burda2019exploration,liang2016shallow}. What we add is a measurement of its effect on the units of the next layer.

Table \ref{tab:adds} summarises this section and the findings of the paper.

\begingroup\footnotesize\setlength{\tabcolsep}{3pt}
\begin{longtable}{>{\raggedright\arraybackslash}p{0.3000\dimexpr\linewidth-6\tabcolsep\relax}>{\raggedright\arraybackslash}p{0.2000\dimexpr\linewidth-6\tabcolsep\relax}>{\raggedright\arraybackslash}p{0.5000\dimexpr\linewidth-6\tabcolsep\relax}}
\caption{What was known, what was missing, and what this paper found. The middle column lists what we did not find in the earlier work we searched, and our search cannot rule out that some of it exists. Each finding carries the limits given in the section that reports it.}\label{tab:adds}\\
\toprule
what was known & what was missing & what this paper found \\
\midrule\endfirsthead
\toprule
what was known & what was missing & what this paper found \\
\midrule\endhead
\bottomrule\endfoot
When the inputs to a layer of ReLUs are far from zero, many units are dead at the start of training, and subtracting the mean of the inputs largely prevents this. Shown for sparse autoencoders \citep{simon2026outliers}. Centring the inputs of a layer is older advice \citep{lecun1998efficient} & Whether a large offset also makes units die during training, beyond one analysis of weight drift \citep{shvetsov2026bugfeature}; whether changing the offset changes the number of live units in other kinds of network, such as a reinforcement-learning agent & A higher offset left fewer units alive at the end of training in each kind of system we tried: agents on three games, image classifiers, a reservoir computer and sparse autoencoders (Table \ref{tab:settings}). In sparse autoencoders on GPT-2 at four and eight times the unaltered offset, most of the units that were dead at the end had been alive at the start (learning rate 0.001). At a learning rate of 0.003, subtracting the average lowered the fraction of dead units from 0.407 to 0.005. Section 4.3 gives a caution about the sparsity penalty in these tests. In Pong, subtracting the average screen raised the number of live units from between 3 and 6 of 64 to between 59 and 63, by the 1 percent rule \\
\addlinespace[6pt]
The number of dead or dormant units is used as a sign of lost plasticity \citep{sokar2023dormant,lyle2023plasticity,dohare2024plasticity}. In widened agents a large share of the dormant units are in the dense layer after the encoder \citep{sokar2025gap}. Our earlier study took the number of live units as a measure of the game's complexity \citep{norton2026frozen} & A tested cause for the dead units in the layer after the encoder; what the number of live units depends on; whether raising it on purpose improves learning & In agents with a frozen encoder the number depended on the offset and, less firmly, on the step size. Both can be measured before training. The same game with the same encoders ended with between 3 and 6 or between 59 and 63 live units, depending on the offset, so the order of the counts across games does not show how complex the games are. A smaller learning rate for one layer also raised the number. Between 36 and 46 units stayed alive in Breakout, against 9 to 26 (1 percent rule), but those agents scored lower, and the smaller learning rate may itself have slowed learning \\
\addlinespace[6pt]
Under Adam one step changes each weight by at most about the learning rate \citep{kingma2015adam}. Early weight changes line up with the inputs \citep{kosson2026weightdecay}. The learning rate should be reduced in proportion to the fan-in \citep{yang2021tensor}. The learning rate affects how many units die \citep{dufortlabbe2025demp}, and the layer with the most inputs has the most dormant units \citep{sokar2025gap} & An account of that pattern in terms of the learning rate multiplied by the fan-in; a test of how the learning rate that keeps units alive changes with the fan-in & In Breakout agents with unaltered frozen encoders, the layer whose units have 3,136 inputs lost a median of 0.750 of its units and the layer whose units have 64 inputs lost 0.266, by the 0.01 rule. The order was the same under most counting rules. In image classifiers, the learning rate that keeps the number of live units fixed fell a little faster than in proportion to the fan-in (an exponent of 1.158), or roughly in proportion without the largest fan-in (0.923). In sparse autoencoders the first test of this failed, and a test corrected afterwards gave an exponent of 0.971. This evidence is weaker than the evidence for the offset \\
\addlinespace[6pt]
Frozen random networks are used as baselines, as signals for exploration and as front ends of controllers \citep{burda2019exploration,burda2019largescale,seo2021state,chang2020reinforcement}. The outputs of networks with random weights have been analysed mathematically \citep{cho2009kernel,garrigaalonso2019deep,novak2019bayesian} & How the offset of a random encoder's output relates to the offset of its input, and a formula for it that uses only the inputs and the layer sizes & The output of a random encoder had roughly the same offset as its input, between 0.65 and 1.35 times it wherever the pixel offset was at least 1.5. So a game whose screen is bright and mostly unchanging gives the next layer a large offset. A formula that uses only the inputs and the layer sizes predicted the median offset within 25 percent in all 44 conditions, 33 of them fixed in advance \\
\addlinespace[6pt]
Frozen random encoders did much worse than trained ones in four Atari games \citep{stooke2021decoupling}. They can reach expert play in Pong and, without sticky actions, in Breakout, with much longer training \citep{norton2026frozen} & Whether lowering the offset closes the gap between a frozen and a trained encoder & With the average screen subtracted, a frozen encoder learned Pong about as fast as a trained encoder did on unaltered screens. It needed 0.699 to 1.216 times as many steps to reach a training score of 18, on medians of three runs. On unaltered screens, no agent with a frozen encoder reached 18 in 50 million steps. In Breakout the subtraction did not close the gap. The frozen encoder's average training score over the run stayed 199.736 to 232.404 points below the trained encoder's, with or without it. A trained encoder lowered its own offset early in training \\
\addlinespace[6pt]
Some training pipelines centre or normalise their inputs and others do not. Normalising observations matters for performance \citep{andrychowicz2021matters,lee2025simba}. Centring was found to do little against dead units, with the benefit of layer normalisation coming from rescaling \citep{lyle2024normalization}. Subtracting a background or a running mean is established practice \citep{burda2019exploration,liang2016shallow} & Whether, after a fixed front end, subtracting the mean over the data or normalising keeps more units alive, and where a normalising layer works best & In image classifiers, subtracting the mean over the data kept 170 and 185 more of 256 units alive at the two bias values, more than the three other operations we had set rules for. Layer normalisation of the encoder's outputs helped at only one of the two bias values. Applied to the input sums just before the ReLU, it kept nearly every unit alive, 250 of 256 against 7 at one bias value and 255 against 12 at the other. In Pong, a shift after the encoder that leaves the screens alone gave between 56 and 63 live units with the offset lowered and 2 or 3 with it raised. The classifier results are mixed and do not show which operation does the work in the agents \\
\end{longtable}\endgroup

\section{Setup}

\subsection{Agents with a frozen random encoder}

Most of our experiments use agents that learn to play \textbf{Atari} games, a standard benchmark in which an agent sees the screen of the game console and has to learn from the score which buttons to press \citep{bellemare2013arcade}. We use three games, Breakout, Pong and Space Invaders.

\begin{figure}[htbp]
\centering
\includegraphics[width=\linewidth,height=0.80\textheight,keepaspectratio]{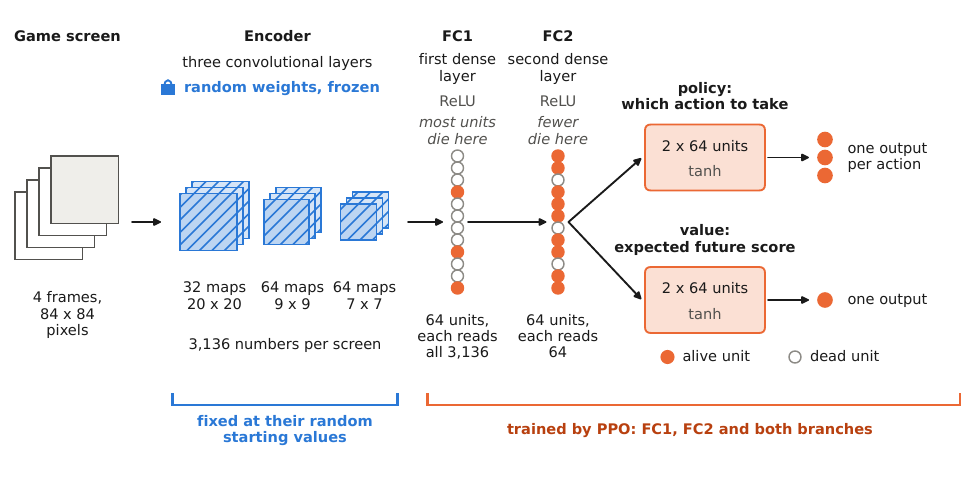}
{\captionsetup{font=small}\caption{The network of our agents. Four stacked greyscale frames of 84 by 84 pixels pass through an encoder of three convolutional layers whose random weights are never trained, which gives 3,136 numbers for each screen. These feed the first dense layer (FC1) and then the second (FC2), of 64 units each, and after FC2 a policy branch and a value branch, each with two more layers of 64 units. Everything after the encoder is trained. The filled and open circles illustrate alive and dead units and are not counts.}\label{fig:network}}
\end{figure}

Figure \ref{fig:network} shows the network. Its input is four greyscale frames of the game, stacked together. The first three layers are convolutional, and together we call them the \textbf{encoder}. For each screen the encoder produces 3,136 numbers. In our agents the encoder keeps its random starting weights and is never trained, and we call it a \textbf{frozen random encoder}. After the encoder come two dense layers of 64 ReLU units each. We call them the first dense layer, or \textbf{FC1}, and the second dense layer, or \textbf{FC2}. FC1 is the first trained layer, and it is the layer this paper is mostly about. After FC2 the network divides into two branches. The policy branch chooses the action, and the value branch estimates how much reward is still to come. In some experiments we train the encoder as well, and then we call it a \textbf{trained encoder}.

The agents are trained with proximal policy optimisation, or \textbf{PPO} \citep{schulman2017proximal}. PPO lets the agent play for a stretch and then adjusts the network so that the actions that were rewarded become more likely. We use the PPO implementation of the Stable-Baselines3 library \citep{raffin2021stable}, with a learning rate of 2.5e-4.

Three terms are used throughout the paper. An \textbf{agent step} is one action chosen by the agent, which the emulator carries out for four frames. The \textbf{training score} is the score of the games that the agent plays while it is learning. \textbf{Sticky actions} are a feature of the benchmark in which the game, with some probability, repeats the agent's previous action in place of the new one. Some of our runs fix that probability at 0.25, and others switch sticky actions off. Unless we say otherwise, a run lasts 50 million agent steps. The expert-level scores that we quote come from the much longer runs of the earlier study \citep{norton2026frozen} (Section 6.1).

\subsection{The offset, and how we change it}

The offset is defined as follows. Take a set of screens, pass each one through the encoder, and write the results as the rows of a table $F$, one row for each screen. Each row is one of the points of the picture in the Introduction, and the table is the collection of points. The \textbf{offset} is $\gamma(F)=\lVert\operatorname{mean}(F)\rVert/\sqrt{\operatorname{tr}\operatorname{cov}(F)}$. The top of this fraction is the length of the average row, and it says how far the centre of the rows is from zero. The bottom is the square root of the sum of the variances of the outputs. We call it the \textbf{spread}, and it says how much the rows differ from one another. The same ratio appears in the work of Simon, Adams and Zou \citeyearpar{simon2026outliers}, who call it the outlier-severity ratio.

An offset is always measured on a particular \textbf{set of screens}, which is a fixed collection of screens that we saved in advance. Two examples are the 50,000 random-play Breakout screens, which were recorded while the game was played with random actions, and the mixed set, which adds 50,000 screens recorded during expert play (Supplement S8.4). The choice of set matters a great deal, because offsets differ by a factor of 3 to 8 from one set to another. For example, the six unaltered Breakout encoders have offsets of 2.787 to 3.061 on the mixed set and of 13.005 to 18.459 on the random-play screens. For this reason we always say which screens an offset was measured on. Sections 4 and 5 give offsets measured on the mixed sets or, for Pong, on a set of 50,000 saved Pong screens. Section 6 gives offsets at several points during training, measured on random-play screens.

We change the offset in three ways. To raise it, we add a constant to the bias of the encoder's last convolutional layer. We call an encoder changed in this way a raised-offset encoder, and an encoder used exactly as it was randomly drawn an unaltered encoder. To lower the offset, we \textbf{centre} the input. That is, we compute the average screen of a saved set of screens, and we subtract it from every screen before the encoder sees it. We call the result \textbf{centred screens}, and we call the screens as the game produces them \textbf{unaltered screens}. The third way is a \textbf{shift after the encoder}. Here we add one fixed list of 3,136 numbers to the encoder's output on every screen. This changes the offset without changing the screens.

\subsection{Counting a unit as alive, and checking that the count matters}

A unit is dead on a set of screens if its input sum is not positive on any of them. To count the units that are alive, we use one of two rules. Under the \textbf{0.01 rule}, a unit counts as alive if its average output over the screens is above 0.01, which means that its output is large enough to matter. Under the \textbf{1 percent rule}, a unit counts as alive if its input sum is positive on at least 1 percent of the screens, which means that it switches on at least occasionally.

Under either rule, a unit that is not counted as alive can still give a small output on a few screens, so it is not always dead in the strict sense of Section 1. We still call such units dead, to keep the wording simple. We treat the 1 percent rule as our main rule. We chose it using an order of preference that we wrote down before the analysis. The two rules usually agree. In 54 Breakout runs they gave the same number in 48, and the six runs in which they differed were the six centred agents, where the difference was at most 10 units. Some experiments report the 0.01 rule, because we wrote their pass and fail levels for that rule before running them. We say which rule and which screens each count uses, and counts made with different rules or on different screens should not be compared with one another. Supplement S17.1 compares fifteen counting rules of this kind.

The number of live units means little unless the units we count as alive are the ones the agent actually uses. The \textbf{silencing check} tests this. We take a trained agent, set the outputs of some of its units to zero, and let it play again. If we silence the units that we counted as alive, the score should fall to a low level that we fixed in advance. If we silence the units that we counted as dead, the score should stay about the same. The check passed in the first Breakout experiments, and it only just passed in Space Invaders. It failed in three experiments. These were one offset level of the Breakout follow-up experiment in Section 4.2, an older experiment with a trained encoder (Section 6.4), and an experiment on the width of FC1 that we report only in the supplement. We did not run the check on the Pong agents of Section 4.2, on the repeated runs of Section 3.5, or on the agents of Sections 6 and 7. Wherever the check failed or was not run, we have not shown that the units counted as alive are the ones the agent depends on (Supplement S5.3 and S17.4).

\subsection{Settings outside games}

We also ran experiments in three kinds of system that have nothing to do with games.

The \textbf{image classifiers} are small networks that classify photographs from CIFAR-10 \citep{krizhevsky2009learning}. Like the agents, they have a frozen random convolutional encoder followed by a trained hidden layer of ReLU units. The hidden layer has 256 units in the experiments on fan-in. Some classifier experiments use synthetic images, and we say so where that is the case.

The \textbf{reservoir computer} is a recurrent network with fixed random weights. It is driven by the Lorenz 96 system, which is a standard chaotic test system, and its state is passed to a trained layer of 64 hidden units.

The \textbf{sparse autoencoders} are trained on the activations of the language model GPT-2 small \citep{radford2019language}, taken at the input of its layer 8. We load the model from its public weights, and the autoencoder is our own implementation. Each autoencoder has eight times as many units as it has inputs, and it is trained for 20,000 steps of Adam (Supplement S6.1).

In each of these systems the picture of the Introduction applies with a different kind of example. For the image classifiers an example is one image, and its point is the output of the encoder for that image. For the reservoir computer it is the state of the reservoir at one moment. For the sparse autoencoders it is the list of 768 numbers that GPT-2 holds at one position in a text.

\subsection{Pre-set tests and the difference between repeated runs}

Two points about our method apply throughout the paper.

The first is that for most experiments we fixed our criteria in advance. Before the experiment we wrote down what result would count as a pass and what result would count as a fail. Where an experiment or a comparison had no such rule, we say so. We call this a \textbf{pre-set test}. Supplement S5 lists every pre-set test with its outcome, including the ones that failed.

The second point is about repeated runs. Training an agent involves a good deal of randomness, so two runs with identical settings do not give identical results. Even so, most of our comparisons between agents use one run for each encoder in each setting. To find out how far a single run can be trusted, we trained each of six unaltered Breakout encoders three times, with a frozen encoder, unaltered screens and sticky actions off. By the 1 percent rule, the three runs of the same encoder differed by up to 7 live units, and their average training score over the run differed by up to 16.45 points. We use these two figures as a reference. When two single Breakout runs differ by 7 units or fewer, the difference is within what repeated runs alone can produce. Three runs is a small sample, so the true range could be larger. In Pong we made three runs per encoder in two settings, over the first 10 million agent steps. Those runs differed by a factor of up to 1.796 in the agent step at which the training score first reached 18, and by up to 4.063 points in the average training score (Supplement S4.7).

One more term is needed for the tables. A rank correlation measures how consistently one quantity rises or falls as another rises, on a scale from $-1$ to $+1$.

\section{The number of live units depends on the offset and the step size}

This section presents the first finding. The number of units that are alive at the end of training depends on two quantities that are known before training starts, the offset of the layer's inputs and the step size. The evidence is stronger for the offset than for the step size. In the agents and the image classifiers, the changes we made to the offset also changed the size of the inputs, and Section 8.3 says what follows from that. Earlier work had tied the offset to units that are dead at the start of training in sparse autoencoders. The experiments here show that it also affects how many units are alive at the end of training, in four kinds of system. They also show that we can move the number up or down by changing the offset on purpose. Table \ref{tab:settings} lists every system in which we raised or lowered the offset, and in each of them a higher offset left fewer units alive, although not every pre-set test of the effect passed. We go through these experiments first, then turn to the step size, and end with the effect of the task.

\begin{table}[htbp]
\centering
\caption{Settings in which the offset was raised or lowered, with the system each result comes from. Counts for the agents are of FC1 units alive. The Breakout counts use the 0.01 rule, and the Pong and Space Invaders counts use the 1 percent rule. The image-classifier figures are medians over a sweep on synthetic images. Counts in different systems are not comparable. Details are in Supplement S10 (agents, image classifiers and the reservoir computer) and in Supplement S16.3 and S6.2 (sparse autoencoders).}\label{tab:settings}
{\scriptsize\setlength{\tabcolsep}{3pt}
\begin{tabular}{>{\raggedright\arraybackslash}p{0.1100\dimexpr\linewidth-14\tabcolsep\relax}>{\raggedright\arraybackslash}p{0.1600\dimexpr\linewidth-14\tabcolsep\relax}>{\raggedright\arraybackslash}p{0.1700\dimexpr\linewidth-14\tabcolsep\relax}>{\raggedright\arraybackslash}p{0.1500\dimexpr\linewidth-14\tabcolsep\relax}>{\raggedright\arraybackslash}p{0.1500\dimexpr\linewidth-14\tabcolsep\relax}>{\raggedright\arraybackslash}p{0.1100\dimexpr\linewidth-14\tabcolsep\relax}>{\raggedright\arraybackslash}p{0.1500\dimexpr\linewidth-14\tabcolsep\relax}}
\toprule
setting & system & how the offset was changed & at the lower offset & at the higher offset & runs & pre-set test \\
\midrule
Breakout agents & PPO, Stable-Baselines3 & a constant added to the encoder's last bias; the average screen subtracted & unaltered: median 16 (15 to 21) of 64; centred: median 27 & aimed at 5, 8, 12 and 16: medians 2.0, 1.5, 0.0 and 1.0 & 30 agents (6 unaltered, 24 raised); 6 centred & rank correlation of $-0.8$ or lower: failed ($-0.7474$); single step: failed; centred median of at least 24: passed \\
\addlinespace[4pt]
Space Invaders agents & PPO, Stable-Baselines3 & a constant added to the encoder's last bias & unaltered: at least 37 of 64 & at 2.50: at most 21; at 3.13: at most 20 & 12 agents (3 encoders, 4 offset levels) & the same offset as in Breakout (median at 3.13 of 11 or less): not met (12) \\
\addlinespace[4pt]
Pong agents & PPO, Stable-Baselines3 & the average screen subtracted; a shift after the encoder & centred: 59 to 63 of 64; shift lowering the offset: 56 to 63 & unaltered: 3 to 6; shift raising the offset: 2 or 3 & 6 encoders, one run per setting & rules for the centred count and for both shifts: met \\
\addlinespace[4pt]
image classifiers & supervised networks, frozen random encoder & 2.0 added to the bias of every convolutional layer & median offset 1.077: median 58 of 64 alive & median offset 3.266: median 19 of 64 alive & a sweep over several seeds & no rule on record \\
\addlinespace[4pt]
reservoir computer & fixed random recurrent network, trained layer of 64 units & offset of the features set at five levels & more units alive at lower offsets (Figure \ref{fig:dosecurve}a, one dot per run) & rank correlation $-0.908$ between the offset and the number alive & see Figure \ref{fig:dosecurve}a & no rule on record \\
\addlinespace[4pt]
sparse autoencoders on GPT-2 & own implementation, Adam & activations moved along the average activation, size fixed & share dead 0.000 with the average removed & share dead 0.991 at eight times the unaltered offset (learning rate 0.001) & medians of five seeds & the share does not fall as the offset rises, and is higher at eight times than at one: passed; the same rule with a matched penalty: failed at its first step only \\
\bottomrule
\end{tabular}}
\end{table}

\subsection{The three games differ in offset and in the number of live units}

We start with an observation about the games themselves. With a frozen random encoder, the three games end training with very different numbers of live units. Between Pong and Space Invaders they differ by a factor of three or more. The games also differ in offset, and the two go together. The game with the largest offset has the fewest live units. Taken as a median over 24 random encoders, the offset of the encoder's output is 27.54 in Pong, 2.90 in Breakout and 1.34 in Space Invaders. The numbers of live units reported for the same games, out of 64 and all by the 0.01 rule, are 1 to 11, 15 to 21 and 35 to 44 (Table \ref{tab:games} in Section 5.1). The Pong numbers come from the earlier study, and they cover its runs with and without sticky actions. The runs without sticky actions had 3 to 8.

The earlier study \citep{norton2026frozen} took the order of these counts to be the order of the games' complexity, with a more complex game keeping more units alive. The comparison above suggests a different explanation, which is that the games differ in offset. It cannot settle the question, because the counts and the offsets come from different runs and different screens, and because the three games differ in many ways at once. To test whether the offset is the cause, we have to change the offset and leave everything else as it is.

\subsection{Changing the offset changes the number of live units}

Our first experiment raised the offset in Breakout. We took six unaltered encoders and made four raised-offset versions of each, with target offsets of 5, 8, 12 and 16. This gave 30 agents. Every one of the 6 unaltered agents ended training with more live units than every one of the 24 raised-offset agents. That is 144 comparisons out of 144, although they are not independent of one another, because the agents share encoders. By the 0.01 rule, the unaltered agents ended with 15 to 21 live units, with a median of 16. At the four raised levels the medians were 2.0, 1.5, 0.0 and 1.0. The agents with fewer live units also played worse. The median of the average training score over the run was 21.4 for the unaltered agents, and it was 2.1, 2.0, 1.9 and 2.0 at the four raised levels. The closest of the 144 comparisons was 15 live units against 10. That difference is smaller than the 7 units by which repeated runs of one encoder can differ (Section 3.5). The 7 units were measured with the 1 percent rule, but for unaltered and raised-offset agents the two rules give the same counts (Section 3.3).

\begin{figure}[htbp]
\centering
\includegraphics[width=\textwidth,height=.85\textheight,keepaspectratio]{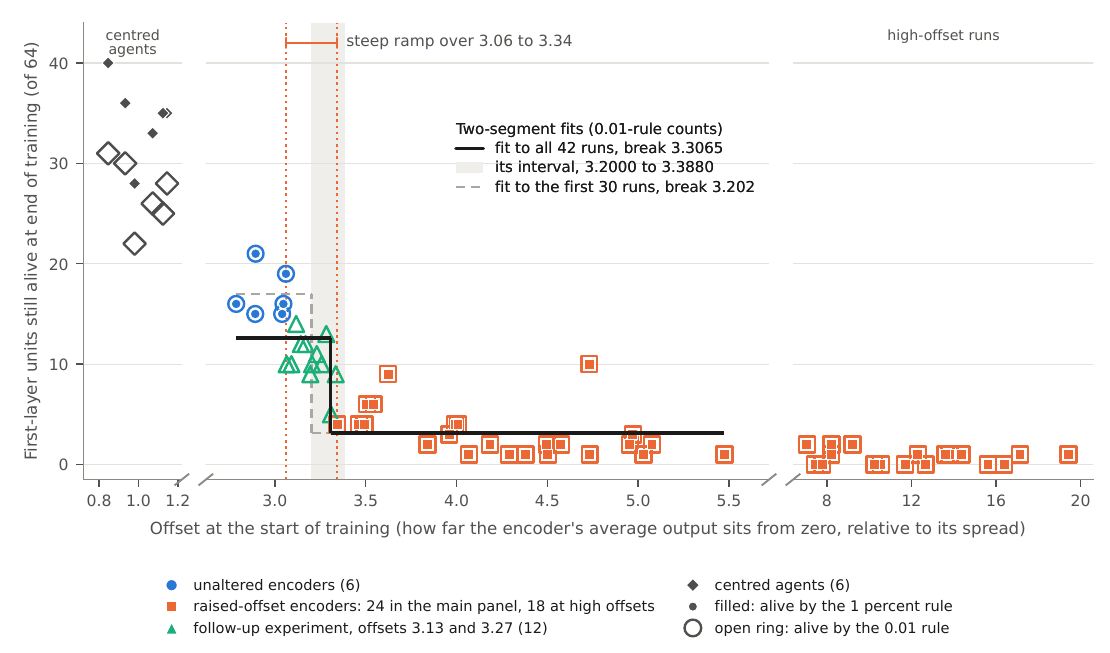}
{\captionsetup{font=small}\caption{Final FC1 units alive in each Breakout agent against its encoder's offset before training. Blue circles are unaltered encoders, orange squares raised-offset encoders, green triangles the 12 follow-up runs and grey diamonds centred agents. Filled markers use the 1 percent rule, open ones the 0.01 rule. The black and dashed steps are two-segment fits (Supplement S2.1). The bracket marks the ramp from 3.06 to 3.34.}\label{fig:ramp}}
\end{figure}

Next we looked more closely at the range of offsets in which the number of live units falls. The first experiment had left a gap between the unaltered encoders and the lowest raised level, so we ran two more experiments. One placed 18 raised-offset agents in the gap. The other raised each unaltered encoder slightly, to offsets of 3.13 and 3.27. We had expected the number of live units to drop suddenly at one particular offset, but it did not. As the offset goes from 3.06 to 3.34, the median falls steeply, in several stages. We call this a ramp (Figure \ref{fig:ramp}).

Two of our pre-set tests failed in these experiments. The first belongs to the first experiment. It required a rank correlation of $-0.8$ or lower between the offset and the final number of live units, and the value was $-0.7474$. The second test was for a sudden drop. It required a median of 12 or more live units at an offset of 3.13 and of 6 or fewer at 3.27, and the medians were 11 and 10. So we do not name a single offset at which the fall happens. We also cannot say where its stages lie, because the medians at neighbouring offsets differ by no more than repeated runs of one encoder do.

Space Invaders gave the same result as Breakout when we raised the offset. By the 1 percent rule, the unaltered agent with the fewest live units had 37. That is more than any agent had at an offset of 2.50, where the largest number was 21, or at an offset of 3.13, where the largest was 20. This experiment is smaller than the Breakout experiments. It has three runs at each offset level, and it has no offset above 3.13 (Supplement S10.4).

We then went the other way and lowered the offset, by centring the screens. In Breakout, the centred agents ended with a median of 27 live units, compared with 16 for the unaltered agents, by the 0.01 rule. This passed our pre-set threshold, which was a median of at least 24. We later repeated the centring with six new runs. By the 1 percent rule these agents ended with 30 to 37 live units. For each encoder, that is 11 to 19 more than the highest of its three unaltered runs.

The largest effect was in Pong, which is also the game with the largest offset. We trained six unaltered encoders on Pong twice, once on unaltered screens and once on centred screens, and we changed nothing else. Before the runs we had written down a rule for what would count as an effect, and the result met it. Centring reduced the offsets from between 22.94 and 32.56 to between 0.691 and 0.764, measured before training on the 50,000 saved Pong screens. The number of live units, by the 1 percent rule on the 25,000 random-play screens, went from between 3 and 6 to between 59 and 63. By the 0.01 rule the centred agents had 38 to 53 live units. So with the same game, the same encoders and the same training, nearly all of the units ended alive by the 1 percent rule, where on unaltered screens nearly all had ended dead.

Centring changes the screens that the encoder sees, so it is fair to ask whether the offset or something else about the screens is responsible. To help separate the two we used the shift after the encoder, which changes the offset and leaves the screens as they are. The number of live units changed with the offset in both directions. When the shift lowered the offset, the agents ended with 56 to 63 live units. When it raised the offset, they ended with 2 or 3. The agents at the lower offset also learned the game far faster, and we come back to this in Sections 6.2 and 7. Like centring, the shift also changes the overall size of FC1's inputs, so it does not separate the offset from that (Section 8.3).

These experiments have limits. Each setting has one run per encoder. The agents being compared differ in how well they play as well as in their number of live units. And we did not run the silencing check on the Pong agents (Section 3.3). Supplement S10.3 gives further limits.

\subsection{The same effect outside games}

Nothing in the explanation of Section 1 is specific to games, so we tested whether the same effect appears in the three systems outside games that Section 3.4 described. In all three, a higher offset left fewer units alive.

\begin{figure}[htbp]
\centering
\includegraphics[width=.92\linewidth,height=0.80\textheight,keepaspectratio]{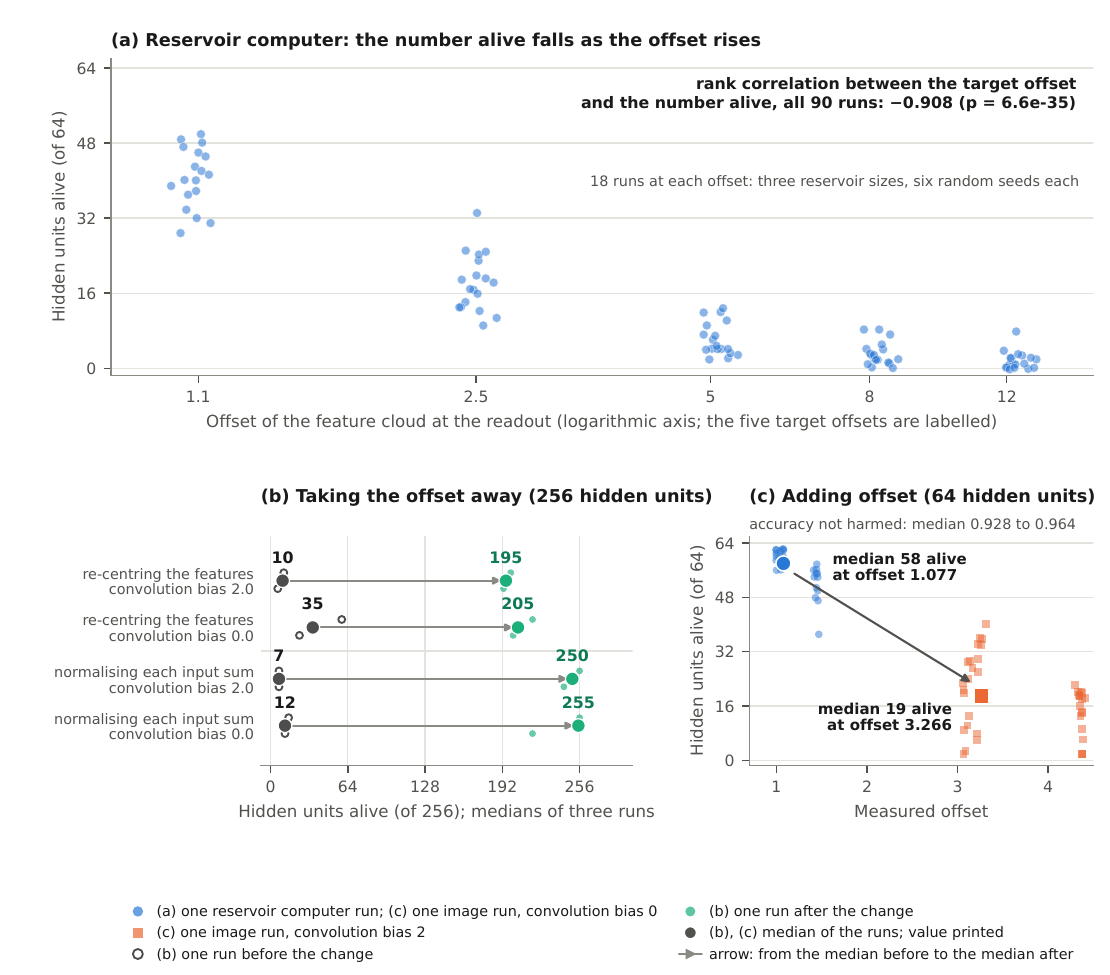}
{\captionsetup{font=footnotesize}\caption{Moving the offset outside the Atari agents. (a) Reservoir computer, one dot per run, on a logarithmic offset axis. (b) Image classifiers with 256 hidden units, median count before (dark) and after (green) subtracting the training-set mean feature vector (upper rows) or applying layer normalisation to the hidden layer's input sums just before the ReLU (lower rows). This placement differs from the one in Table \xlabel{S5}{tab:norm} in Supplement S2.4, which normalises the encoder's outputs. (c) Image classifiers, 64 hidden units, with the bias raised from 0 (blue) to 2 (orange).}\label{fig:dosecurve}}
\end{figure}

In the reservoir computer, we set the offset of the features that the trained layer receives to five different levels. The higher the offset, the fewer hidden units were alive at the end of training. The rank correlation between the offset and the number of live units was $-0.908$ ($p$ = 6.6e-35) (Figure \ref{fig:dosecurve}a).

In the image classifiers we raised the offset by adding a constant of 2.0 to the bias of every convolutional layer of the encoder. The median offset went from 1.077 to 3.266, and the median number of live units went from 58 to 19 out of 64. Accuracy did not fall. The median accuracy in fact rose, from 0.928 to 0.964. These figures are medians pooled over a sweep of runs on synthetic images, so the accuracies should not be compared with accuracies on CIFAR-10. We also lowered the offset. In classifiers with 256 hidden units, at a bias of 2.0, subtracting the mean feature vector of the training set raised the median number of live units over three seeds from 10 to 195 out of 256.

Subtracting a mean and normalising are different operations, so we also asked which operation is responsible for keeping units alive. We tried four operations on the outputs of the encoder under rules set in advance, with three seeds each. Of these, subtracting the mean feature vector of the training set helped the most. It added 170 live units at a bias of 0.0 and 185 at a bias of 2.0. A fifth operation, whitening, for which we had set no rule, added more still, 209 and 230. Two other operations, dividing each feature by its spread and applying layer normalisation to the encoder's outputs, each helped at only one of the two bias values (Supplement S2.4). Where the normalising layer is placed also matters. In a separate set of runs we applied layer normalisation to the input sums of the hidden layer, immediately before the ReLU. The median number of live units over three seeds then went from 7 to 250 out of 256 at a bias of 2.0, and from 12 to 255 at a bias of 0.0 (Figure \ref{fig:dosecurve}b). Taken together these results are mixed, and they do not tell us which operation is responsible for the effect in the Atari agents.

The third system is the sparse autoencoder. A sparse autoencoder is a network with one hidden layer. It is trained to rewrite the activations of a language model as a sum of a few items taken from a large dictionary \citep{bricken2023monosemanticity,cunningham2024sparse}. Its hidden layer is a layer of ReLU units whose inputs stay fixed during training, which is the same situation as FC1 after a frozen encoder. Simon, Adams and Zou \citeyearpar{simon2026outliers} showed that the offset predicts how many units are dead at the start of training. They also report that a strong sparsity penalty or a high learning rate can make units die during training, but they do not analyse those later deaths in terms of the offset.

In our experiment, a unit counts as dead if its output is zero on every one of 400,000 held-out tokens. The activations of GPT-2 as they come, which we call unaltered, have an offset of 0.904. To raise the offset, we moved every activation along the direction of the average activation, and we kept the overall size of the inputs the same. With a learning rate of 0.001, the fraction of units that were dead after training rose with the offset at every level. As a median over five seeds, it was 0.000 when the average had been removed, 0.002 at the unaltered offset, 0.228 at twice the unaltered offset, 0.837 at four times and 0.991 at eight times (Supplement S6.2). Most of these units were alive when training began, so they died during training. At four and eight times the unaltered offset they died early, since most of the units that were dead at the end were already dead by step 50. Subtracting the average activation also helped in a case where units were being lost at the unaltered offset. At a learning rate of 0.003, the fraction of dead units was 0.407 at the unaltered offset and 0.005 with the average removed. This was a pre-set test, and it passed (Supplement S5.7).

One part of this experiment needs a caution. Because we kept the overall size of the inputs fixed, raising the offset made the varying part of the inputs smaller, and the sparsity penalty therefore became stronger in relative terms. The penalty, and not the offset, might then have been what made the units die. To check this, we repeated the experiment with a penalty matched to the varying part of the inputs, and again with no penalty at all. In both cases the fraction of dead units still rose overall between one and eight times the unaltered offset. However, a pre-set rule that the fraction must never fall from one level to the next failed by small margins (Supplement S5.7). With no penalty, the fraction of dead units was already high at the unaltered offset, and we have not been able to explain that. The experiment also covers only one language model, one layer and one training recipe. The training runs are short, and they do not use the remedies that widely used recipes include, such as resampling dead units. The later groups of runs were also chosen after we had seen the results of the earlier ones (Supplement S5.7).

\subsection{The step size}

We now turn to the second quantity. As Section 1 explained, one step under Adam \citep{kingma2015adam} changes each weight by at most about the learning rate. Early in training, the changes to a unit's weights tend to line up with its inputs, so that they all push the input sum in the same direction. Kosson et al. \citeyearpar{kosson2026weightdecay} measured this alignment at the start of training, in runs that trained language models. When the changes line up, the input sum moves by about the learning rate times the unit's fan-in. This last step is our own derivation from their equations and from Adam's update rule, and it is not a result that they state. A unit of FC1 has 3,136 inputs and a unit of FC2 has 64. If the reasoning is right, FC1 should lose more of its units than FC2.

That is what we found. In Breakout agents with unaltered frozen encoders, FC1 lost a median of 0.750 of its units and FC2 lost 0.266. These figures use the 0.01 rule on the random-play screens at the end of training. FC1 lost more than FC2 under 12 of the 15 counting rules we compared, including the 1 percent rule. Under a cut-off of 0.1 on the average output the result is mixed, and at cut-offs of 0.3 and 1.0 the order reverses. Two further cautions apply. The result is for frozen encoders. With a trained encoder, FC1 ended with a median of 53 of its 64 units alive and FC2 with 56, by the 0.01 rule. And our explanation is not the only one that predicts the result. Shvetsov et al. \citeyearpar{shvetsov2026bugfeature} give an analysis in which earlier ReLU layers accumulate more negative weight, and it predicts the same order. This measurement cannot tell the two explanations apart.

In the image classifiers we could measure the relation more directly. We varied the fan-in of the first layer and changed only that layer's learning rate. We then estimated an exponent that describes how quickly the learning rate has to fall as the fan-in grows in order to keep the number of live units the same. An exponent of 1 means that the learning rate has to fall in proportion to the fan-in, as our reasoning predicts, and an exponent of 0.5 means in proportion to its square root. Using all the fan-in values, the exponent was 1.158, with a 95\% interval of 1.109 to 1.181. Leaving out the largest fan-in, 2,048, it was 0.923, with an interval of 0.797 to 1.049, which includes 1. In both cases the square-root rule is ruled out. The exact value should be treated with care, for several reasons. If the units are counted earlier, at step 1,000, the exponent is 0.969 and its interval lies below 1. The result comes from a single task and a single layer width. The later part of the fall in the number of live units may be due to weight decay, which steadily pulls the weights towards zero. And we designed this experiment after we had seen the results of earlier designs (Supplement S12.3 and S1.6).

The sparse autoencoders gave a less clear result. If one step can move a unit's input sum by about the learning rate times the fan-in, then the learning rate at which half of the units die should halve each time the fan-in doubles. We had set a pre-set test of this prediction, which required an exponent between 0.7 and 1.3, and it failed, with an exponent of 0.670. We then found a flaw in the design of the experiment. To vary the fan-in we had made smaller versions of the inputs by random projection, but we had left the full-size inputs, of size 768, in GPT-2's own coordinates. This matters because Adam adjusts each weight separately, so the distance that one step can move a unit's input sum depends on how the input is spread across its coordinates. When we also rotated the inputs of size 768 at random, the exponent rose to 0.971. We took this step only after the first test had failed. We also found that the number of units per input matters, that is, the size of the dictionary relative to the size of its input. With two, four or eight units per input, the learning rate at which half of the units die fell roughly in inverse proportion to the fan-in. With more units per input the fall was weaker.

Finally, we tested the step size in agents by changing the learning rate of one layer. We divided the learning rate of FC1 alone by 49. In Breakout, between 36 and 46 of its units then stayed alive, by the 1 percent rule and with one run per encoder. At the usual learning rate, the three runs of each of the six encoders ended with between 9 and 26. This is what the reasoning predicts. Section 7 describes what the change did to learning, which is a separate question. We did not show whether the two quantities compound, that is, whether a large offset makes a large step more harmful.

\subsection{The effect of the task}

The earlier study proposed that the number of live units depends on the task itself. We tested that idea in the image classifiers, where we can make a task harder while keeping the pictures, or at least the offset, the same.

The direction of the effect was different in the tests with a softmax output layer and in the test with a fixed-width output layer, and these tests also differ from one another in other ways. With a softmax output layer, which has one output unit for each class, a harder task left fewer units alive. We saw this in three tests. In the first, some of the training labels were wrong, and here the evidence comes from a follow-up that was exploratory. In the second, four classes had to be split into two groups in a way that was harder to learn. In the third, the task used CIFAR-10 in place of the easier MNIST. The fixed-width output layer has 8 units, and it is trained by squared error towards a fixed random target for each class. With it, more classes left more units alive. The median went from 10 live units with two classes to 50 with 32 classes, at an offset that stayed level. This second result is what the earlier study's view predicts, and we count it as evidence for that view in that setting. A fourth test, in which each image was a sum of k patterns, showed an increase with k in one of its four variants and no increase in the other three. We do not know why the direction differs between the two kinds of output layer.

We also tried to test the idea in agents, by training Breakout agents with and without sticky actions. This test did not settle the question. Sticky actions were meant to make the game harder, but the training score was lower with them in only 4 of 6 pairs of runs, and our rule required 5.

The task does affect the number of live units. In the test with the harder split, the pictures, the encoder and the starting weights were the same in the easy and the hard version, and the hard version ended with fewer live units in 15 of 15 pairs, although not in the version of the test with a constant of 2.0 added to the encoder's biases. But a number that goes down with difficulty under one output layer and up under another is not a consistent measure of how complex a task is. This changes how the counts in the earlier study should be understood. The same game, Pong, ended with very different numbers of live units depending on the offset of its screens (Section 4.2). So the order of the counts across games is not evidence that the games differ in complexity in that order (Supplement S14.1).

\section{A random encoder leaves the offset of its input roughly unchanged}

This section presents the second finding. We explain where the offset of a game's screens comes from, show that a random encoder leaves that offset roughly unchanged, and give a formula that predicts the offset of the encoder's output from its inputs alone. We did not find earlier work that relates the offset after a random encoder to the offset of its input, or that predicts it.

\subsection{Where a game's offset comes from}

The offset can be measured on any set of inputs, including the raw pixels of the screens before they reach the encoder. For the three games, the offsets of the raw pixels are 25.05 in Pong, 3.34 in Breakout and 1.61 in Space Invaders. Table \ref{tab:games} sets these beside the offsets of the encoder's output.

\begin{table}[htbp]
\centering
\caption{What the offset in each game depends on (Atari agents). The first four number columns come from the raw pixels of the mixed sets of screens with exact duplicate screens removed; the squared pixel offset is the energy divided by the product of the next two. The encoder columns use 10,000 of those screens (Supplement S2.7). The counts come from other runs and other screens than the offsets beside them (Table \xlabel{S6}{tab:gamesdetail} in Supplement S2.6). The Pong count covers the earlier study's runs with and without sticky actions, and the runs without them had 3 to 8.}\label{tab:games}
{\footnotesize\setlength{\tabcolsep}{3pt}
\begin{tabular}{>{\raggedright\arraybackslash}p{0.1407\dimexpr\linewidth-16\tabcolsep\relax}>{\raggedright\arraybackslash}p{0.1022\dimexpr\linewidth-16\tabcolsep\relax}>{\raggedright\arraybackslash}p{0.1134\dimexpr\linewidth-16\tabcolsep\relax}>{\raggedright\arraybackslash}p{0.1577\dimexpr\linewidth-16\tabcolsep\relax}>{\raggedright\arraybackslash}p{0.1310\dimexpr\linewidth-16\tabcolsep\relax}>{\raggedright\arraybackslash}p{0.1358\dimexpr\linewidth-16\tabcolsep\relax}>{\raggedright\arraybackslash}p{0.1134\dimexpr\linewidth-16\tabcolsep\relax}>{\raggedright\arraybackslash}p{0.1059\dimexpr\linewidth-16\tabcolsep\relax}}
\toprule
game & offset of the pixels & energy of the average screen & pixels that change (soft count, of 28,224) & variance of a changing pixel & offset of the encoder's output, median of 24 encoders & formula & alive of 64 \\
\midrule
Pong & 25.05 & 5,865 & 1,774 (6.3\%) & 0.0053 & 27.54 & 28.18 & 1 to 11 \\
Breakout & 3.34 & 1,926 & 4,660 (16.5\%) & 0.0371 & 2.90 & 2.91 & 15 to 21 \\
Space Invaders & 1.61 & 249 & 8,896 (31.5\%) & 0.0108 & 1.34 & 1.33 & 35, 40, 44 \\
\bottomrule
\end{tabular}}
\end{table}

\begin{figure}[htbp]
\centering
\includegraphics[width=.88\linewidth,height=0.80\textheight,keepaspectratio]{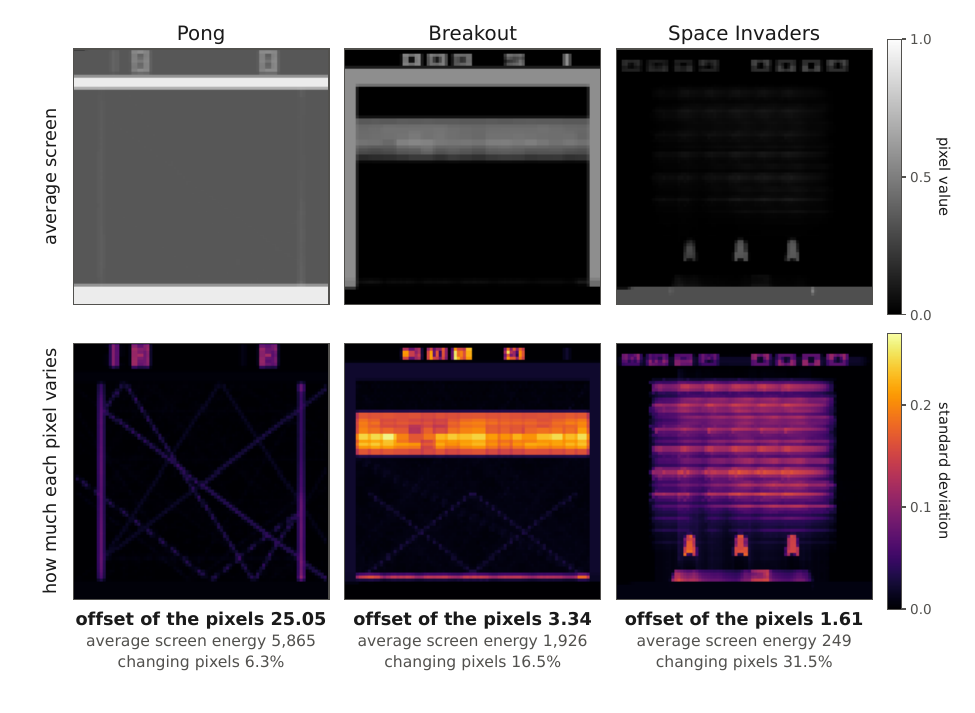}
{\captionsetup{font=small}\caption{The average screen of each game (top) and how much each pixel varies from screen to screen (bottom), with pixel values scaled from 0 to 1. In Pong the thin lines are the paths of the ball. In Breakout the brightest band is the bricks, present in some screens and gone in others.}\label{fig:screens}}
\end{figure}

To see why the games differ, it helps to split each screen into two parts. One part is the average screen, which is what all the screens of a game have in common. The other part is whatever is different about this particular screen (Figure \ref{fig:screens}). A screen is a list of 28,224 pixel values, which is 84 by 84 pixels in each of four stacked frames. Write $m_i$ for value number $i$ of the average screen, and $\sigma_i^2$ for the variance of value number $i$ from one screen to the next. The square of the pixel offset is then $\sum_i m_i^2 \,/\, \sum_i \sigma_i^2$. The top of this fraction is called the \textbf{energy} of the average screen. It is large when the picture that stays the same is bright. The bottom is the sum of the variances. It is large when many pixels change, or when the pixels that change do so by a large amount. So the offset is larger when the unchanging picture is brighter, and smaller when more pixels change.

In Pong the unchanging part of the screen is bright and few pixels change. In Space Invaders the background is dark and many pixels change. To measure how many pixel values change, we use a soft count, the participation ratio of the pixel variances, which gives little weight to pixels that hardly vary. By this count, 6.3 percent of the pixel values change in Pong, 16.5 percent in Breakout and 31.5 percent in Space Invaders. The energy of the average screen differs even more between the games (Table \ref{tab:games}). So Pong and Space Invaders differ mainly in how bright the unchanging picture is, and after that in how much of the screen changes (Supplement S2.6).

The offset takes no account of which pixels matter for playing the game. Any pixel that differs from one screen to the next lowers the offset, whether it shows the ball, the score, or a brick that is present on some screens and gone on others.

\subsection{What the encoder does to the offset}

When the screens pass through a random encoder, the offset stays roughly the same, for the following reason. Each output of a convolutional filter is a sum of pixel values multiplied by random weights, and a random weight is as likely to be positive as negative. On average over the random weights, the square of such a sum is proportional to the energy of whatever the filter receives. That is true whether the filter receives a patch of the average screen or a patch of the differences from the average screen. The filter treats the two parts alike, so on average both keep their energy. Random projections preserve distances for the same reason \citep{johnson1984extensions}.

We checked this in 33 conditions. They consist of the three games, each with its screens as recorded and with its screens changed in ten different ways, for example by making every pixel brighter (Supplement S2.7). Across the 33 conditions, the rank correlation between the offset of the pixels and the offset of the encoder's output was 0.988. Wherever the pixel offset was at least 1.5, the offset of the encoder's output was between 0.65 and 1.35 times the pixel offset.

The match is not exact. Two things shift the offset on the way through the encoder. The first is that neighbouring filter windows overlap, so the pixels near the middle of the screen are covered by many more windows than the pixels near the edge. This lowers the offset when the parts of the screen that hardly change are near the edge, as they are in Space Invaders and Breakout, and it raises the offset a little in Pong (Supplement S2.7). The second is the ReLU at the end of the encoder. A ReLU replaces negative numbers by zero, so its outputs have a positive average even when its inputs average zero. We can see the size of this effect by subtracting the average screen, so that the pixels have no offset at all. Encoders without ReLUs then have an output offset of zero. Encoders with ReLUs still have a median offset of 0.72 in Pong, 0.97 in Breakout and 1.10 in Space Invaders, over 24 encoders in each game. This is consistent with theory for fully connected layers with Gaussian weights \citep{giryes2016deep}. It also means that centring the screens cannot bring the offset of the encoder's output all the way down to zero.

\subsection{A formula that uses only the inputs}

Because the encoder's weights are random, we do not need to build an encoder to know roughly what offset its output will have. We can calculate the offset as an average over the random weights. The calculation uses a known result. For a ReLU unit with random weights, the average value of the product of its outputs on two different inputs can be written down exactly, and it depends only on the lengths of the two inputs and on the angle between them \citep{cho2009kernel}. The same calculation can be repeated for each convolutional layer in turn, one position at a time \citep{garrigaalonso2019deep,novak2019bayesian}. What the formula needs is a sample of screens and the sizes of the layers. No weights are drawn, and nothing is fitted to data. It gives the offset that an encoder of that shape will have, averaged over its random weights (Supplement S2.7).

\begin{figure}[htbp]
\centering
\includegraphics[width=\linewidth,height=0.80\textheight,keepaspectratio]{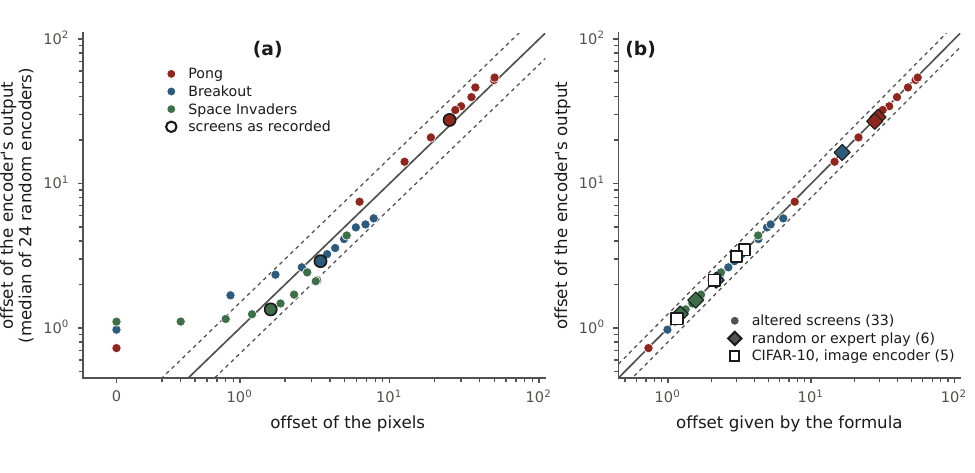}
{\captionsetup{font=small}\caption{The output of a random encoder has roughly the same offset as the pixels, and a formula predicts it. (a) Median encoder offset over 24 random encoders against the offset of the raw pixels, for the 33 conditions of Table \xlabel{S7}{tab:cells} (Supplement S2.7); dashed lines mark a factor of 1.5. (b) The same offsets against the formula, with dashed lines at 25 percent. Diamonds are the random-play and expert-play screens of each game, squares are CIFAR-10 (Tables \xlabel{S8}{tab:natural} and \xlabel{S9}{tab:cifar} in Supplement S2.7).}\label{fig:formula}}
\end{figure}

We tested the formula against the median offset of 24 random encoders, in the same 33 conditions as above. Before the comparison we set a threshold. The formula had to come within 25 percent of the measured median in at least 27 of the 33 conditions (Supplement S5.6). It came within 25 percent in all 33. The median error was 1.3 percent, and the largest error was 11 percent (Figure \ref{fig:formula}).

After seeing these results we added two more checks. For each of them we set the same 25 percent band before running it, and both passed. The first used six sets of screens, which were the random-play screens and the expert-play screens of each game. The formula was within 2.8 percent on all of them. The second check is the one outside games. We passed CIFAR-10 photographs through the encoder of our image classifiers, which has a different shape from the encoder of the agents. The formula was within 4 percent in all five conditions. Together with the first 33, these make the 44 conditions, and over all 44 the median error was 1.4 percent.

The formula has three limits. First, it gives the median over many random encoders, and a single encoder can be quite different. In Pong, the 24 encoders had offsets from 18.76 to 44.37, around a median of 27.54. Second, we tested it only on random weights and on two encoder shapes, and we did not test whether it predicts the number of live units. Third, it does not apply to a trained encoder, because the weights of a trained encoder are no longer random. So, within these limits, candidate front ends can be compared by their offset before any of them is built. We did not find this calculation used elsewhere to predict the offset, although our search cannot rule that out.

A frozen encoder cannot adjust to its inputs. The offset of its output stays as it is, because its weights never change. Burda et al. \citeyearpar{burda2019exploration} make a similar point about the scale of the inputs. They find that normalising the observations is essential when the target is a random network. Supplement S11 and S2.7 give the details of this section.

\section{A frozen encoder compared with a trained one}

This section presents the third finding. Here we look at how well the agents learn, and not only at how many of their units stay alive. We first recall that a frozen random encoder can reach expert play, and how long that takes. We then compare frozen and trained encoders that start from the same weights. The question is whether lowering the offset removes the difference between a frozen and a trained encoder. The result was different in the two games.

\subsection{Expert play with a random encoder, and how long it takes}

In the earlier study \citep{norton2026frozen}, agents with a frozen random encoder reached expert-level play in Pong and, with sticky actions turned off, in Breakout. In Pong, all five agents finished between +16.3 and +20.1, in a game where the score runs from $-21$ to $+21$. In Breakout, with sticky actions off, three agents scored 221, 351 and 303 after 400, 777 and 853 million agent steps. For comparison, one agent with a trained encoder scored 388.

These agents ended with few live units in FC1. There were between 3 and 8 of the 64 in Pong, and between 20 and 23 in Breakout. These counts use the 0.01 rule, and they were made on screens from each agent's own games. Screens of that kind give higher counts than the random-play screens that we use elsewhere. So few live units did not prevent expert play. In these two games, a network that ended with few live units was still a successful one, once it had been trained for long enough.

The drawback is the training time. Even with sticky actions off, the Pong agents with a frozen encoder needed at least 8.9 times as many agent steps as an agent with a trained encoder before their average score reached zero. The Breakout agents needed 400 to 853 million agent steps. The Pong agents with a trained encoder, by comparison, scored +20.7 after 10 million agent steps.

In Breakout, expert play was reached only with sticky actions off. With sticky actions fixed at a probability of 0.25, two much shorter Breakout runs stayed at about 3 points. In Pong, the one run with sticky actions fixed at 0.25 took 318.6 million agent steps before its training score rose above zero, against 44.6 to 124.5 million for the five runs without them. Three other Pong runs, in which sticky actions were brought in gradually, scored +14, +18 and +19 after much longer training. All of these runs differ from the runs without sticky actions in more than the sticky-action setting, so the comparisons are not clean ones (Supplement S9).

\subsection{The same networks in four settings: Pong}

To compare a frozen encoder with a trained one fairly, we ran an experiment in which everything else was kept the same. We took the six unaltered encoders of Section 4.2. For each one we trained the same network in four settings, with the encoder frozen or trained, and on unaltered or centred screens (Section 3.2). In all four settings the network started from the same weights. We did this in Pong and in Breakout, for 50 million agent steps with sticky actions off (Table \ref{tab:samestart}). A trained encoder here means our usual trainer with the encoder allowed to train and nothing else changed, so the trainer was not tuned for a trained encoder.

\begin{table}[htbp]
\centering
\caption{The same six networks in four settings, after 50 million agent steps with sticky actions off. Each cell gives the lowest and the highest of the six encoders. Steps are the agent step, in millions, at which the training score first reached 18 in Pong or 10 in Breakout. Units alive, of 64, are by the 1 percent rule, and offsets are those at 50 million steps, which for a frozen encoder are its offsets throughout. Both are measured with each model's own encoder on random-play screens, the first 25,000 Pong screens or the 50,000 Breakout screens. Each setting has one run of 50 million steps per encoder, except the frozen encoder on unaltered Breakout screens, which has three. For it the score and the step are medians of the three runs, and the count spans all 18. Supplement S4.6 gives every encoder's values.}\label{tab:samestart}
{\footnotesize\setlength{\tabcolsep}{3pt}
\begin{tabular}{>{\raggedright\arraybackslash}p{0.1486\dimexpr\linewidth-16\tabcolsep\relax}>{\raggedright\arraybackslash}p{0.1408\dimexpr\linewidth-16\tabcolsep\relax}>{\raggedright\arraybackslash}p{0.0788\dimexpr\linewidth-16\tabcolsep\relax}>{\raggedright\arraybackslash}p{0.0982\dimexpr\linewidth-16\tabcolsep\relax}>{\raggedright\arraybackslash}p{0.1347\dimexpr\linewidth-16\tabcolsep\relax}>{\raggedright\arraybackslash}p{0.1411\dimexpr\linewidth-16\tabcolsep\relax}>{\raggedright\arraybackslash}p{0.1255\dimexpr\linewidth-16\tabcolsep\relax}>{\raggedright\arraybackslash}p{0.1323\dimexpr\linewidth-16\tabcolsep\relax}}
\toprule
setting & Pong: steps to a score of 18 (millions) & Pong: units alive & Pong: offset & Breakout: average training score & Breakout: steps to a score of 10 (millions) & Breakout: units alive & Breakout: offset \\
\midrule
frozen encoder, unaltered screens & not reached & 3 to 6 & 23.94 to 34.28 & 15.453 to 26.984 & 4.92 to 7.03 & 9 to 26 & 13.005 to 18.459 \\
\addlinespace[3pt]
frozen encoder, centred screens & 3.26 to 4.62 & 59 to 63 & 0.721 to 0.793 & 21.487 to 24.345 & 1.18 to 2.13 & 30 to 37 & 2.921 to 4.245 \\
\addlinespace[3pt]
trained encoder, unaltered screens & 2.83 to 4.64 & 63 to 64 & 0.395 to 0.431 & 218.743 to 240.696 & 0.70 to 0.84 & 43 to 55 & 1.019 to 1.712 \\
\addlinespace[3pt]
trained encoder, centred screens & 1.67 to 2.29 & 64 & 0.283 to 0.366 & 226.279 to 254.003 & 0.54 to 0.70 & 42 to 53 & 1.172 to 2.269 \\
\bottomrule
\end{tabular}}
\end{table}

\begin{figure}[htbp]
\centering
\includegraphics[width=\textwidth,height=.85\textheight,keepaspectratio]{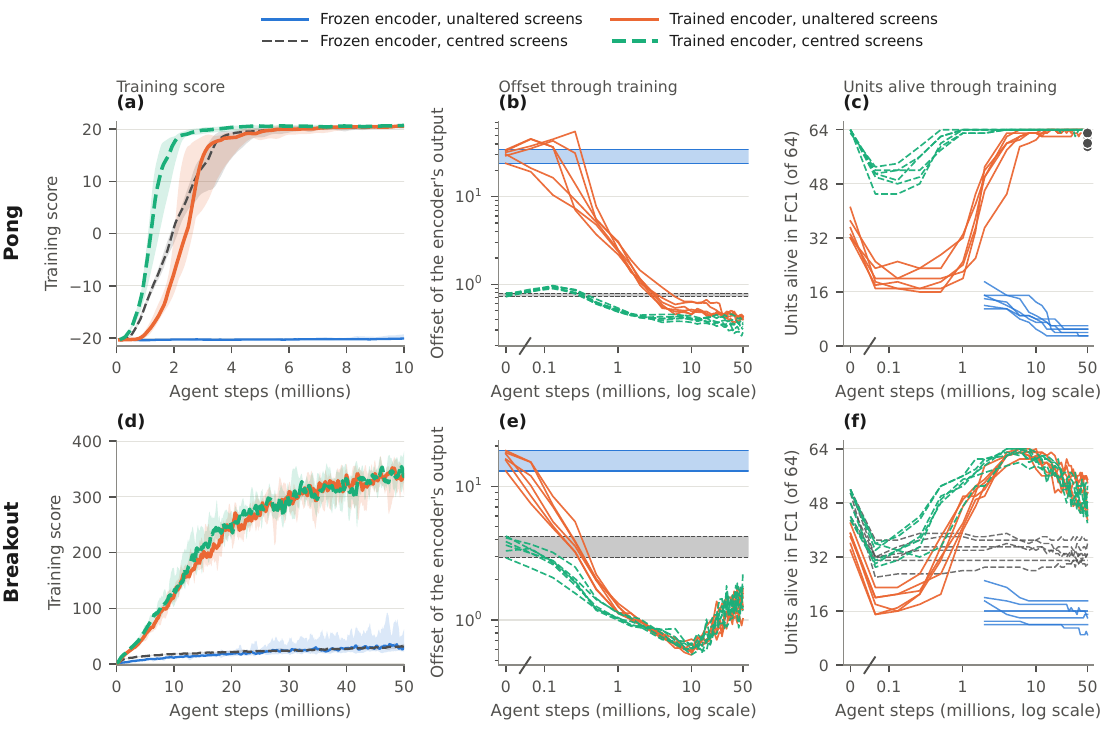}
{\captionsetup{font=small}\caption{The same six networks in four settings, in Pong (top) and Breakout (bottom): encoder frozen or trained, screens unaltered or centred (dashed). Left: training score, the median of the six agents with a band from the lowest to the highest, each curve smoothed over 25 logged points. Middle: the offset of the encoder's output on random-play screens. Lines are trained encoders and bands the range of the frozen ones, which do not change. Right: FC1 units alive by the 1 percent rule, one line per agent. The dots in the top right panel are the frozen-encoder agents on centred screens, which were counted only at the end.}\label{fig:samestart}}
\end{figure}

We begin with Pong. With a frozen encoder and unaltered screens, no agent reached a training score of 18 in 50 million agent steps. With a frozen encoder and centred screens, the agents reached 18 about as early as the agents with a trained encoder did on unaltered screens (Table \ref{tab:samestart}, Figure \ref{fig:samestart}).

We had written down in advance what would count as ``about as early''. For each encoder, we take the agent step at which the frozen encoder on centred screens first reached 18, and we divide it by the agent step at which the trained encoder on unaltered screens first reached 18. The rule required this ratio to be 1.5 or less in at least 5 of the 6 encoders. With one run per encoder and setting, the rule was met in 5. We then made two more runs of each encoder in both settings, over the first 10 million agent steps and under rules that we fixed before they started. Using the median of the three runs, the ratio was between 0.699 and 1.216, and it was 1.5 or less in all six encoders. In all 18 runs on centred screens, the first runs and the repeats together, the frozen encoder reached 18 between 2.54 and 4.92 million agent steps. We should add that the first frozen-encoder runs of this comparison were made with an earlier version of the trainer, and that we checked that version against the newer one in short test runs only.

The repeated runs show that ``about as early'' cannot be taken strictly. Three runs of the same encoder in the same setting differed from one another by a factor of up to 1.796 in the step at which they reached 18. That is more than the 1.5 that the rule allowed. So when we say that the frozen encoder learned Pong ``about as fast'', we mean that the two settings did not differ by more than repeated runs of a single setting differ.

In single runs, the fastest setting was the trained encoder on centred screens. The frozen encoder on centred screens needed 1.421 to 2.274 times as many agent steps to reach 18 as the trained encoder on centred screens did. We had set no rule for this comparison, and we made no repeated runs for it. The smaller of these ratios are within the factor by which repeated runs differ.

We also checked directly whether the output of a frozen encoder on unaltered screens still holds the positions of the ball and the paddles. For each of the six encoders we fitted a straight line, that is, a weighted sum of the encoder's 3,136 outputs, to the position of the ball and to the heights of the two paddles. Each fit was made on one half of the random-play screens and scored on the other half by its R squared, which is 1 for a fit that gives every position exactly and 0 for one that does no better than the average position. R squared was between 0.832 and 0.985 on unaltered screens, and between 0.935 and 0.985 on centred screens (Table \xlabel{S62}{tab:readout} in Supplement S4.8). So a straight-line fit could still read these positions from the encoder's output on unaltered screens, which is consistent with our account that the slow learning there comes from the layer after the encoder and not from the encoder itself. The check was added after the experiments, it is not one of our pre-set tests, and it does not test that account directly.

\subsection{Breakout}

In Breakout the result was different. Centring did not bring the frozen encoder close to the trained one. Over the whole run, the average training score of the trained encoder was 199.736 to 221.008 points higher than that of the frozen encoder on unaltered screens, and 204.292 to 232.404 points higher on centred screens. For comparison, repeated runs differed by up to 16.45 points.

We could not show that centring changed the scores with either kind of encoder. With a frozen encoder, it changed the average training score by less than 16.45 points in all six encoders, although the agents did reach a first score of 10 sooner, by a factor of 3.300 to 4.895. With a trained encoder, the change was larger than 16.45 points in two of the six encoders (+27.283 and +24.349) and negative in two others. By the rule we had written down before the runs, this counts as no difference shown.

We do not know why Breakout differs from Pong. One fact may be relevant. On random-play screens, centring leaves a frozen Breakout encoder with an offset of 2.921 to 4.245, and in Pong it leaves an offset of 0.721 to 0.793. These values are measured on random-play screens, and they should not be compared with the offsets of Section 4.2, which were measured on the mixed set of screens. In Breakout, then, a good deal of the offset remains after centring. These runs therefore cannot tell us whether the remaining offset or some other property of a random encoder is responsible for the lower scores. We did not test whether removing the rest of the offset would close the gap.

\subsection{A trained encoder lowered its own offset}

The same experiment shows what a trained encoder did when it was given unaltered screens, with their large offset. In all twelve runs it lowered the offset by itself (Figure \ref{fig:samestart}, middle). Measured on random-play screens, the offset of the encoder's output in Pong fell from between 23.94 and 34.28 at the start to between 0.395 and 0.431 at the end. In Breakout it fell from between 13.005 and 18.459 to between 1.019 and 1.712. The fall happened early. In all twelve runs the offset first went below 2.0 within the first 1,507,328 agent steps.

The number of live units in FC1 changed over the same period. FC1 first lost units and later regained them. Within the first 507,904 agent steps, the number of live units fell to between 16 and 23 in Pong and to between 15 and 23 in Breakout. At 50 million agent steps it was 63 or 64 in Pong. In Breakout it was between 43 and 55, having declined from an earlier peak while the offset rose again. We did not test what caused that later decline. Before the runs we had written down two rules. One was that the offset would be below 2.0 at 50 million steps. The other was that more units would be alive than the frozen encoder kept on unaltered screens. Both were met in all 6 encoders, in both games.

It is tempting to conclude that the units come back because the offset falls. Our data do not show that. They show that the two things happen at about the same time, and we did not test whether one causes the other.

Sokar and Castro \citeyearpar{sokar2025gap} found that, in widened agents whose encoders are trained, a large share of the dormant units are in the first dense layer. We did not see this in our agents. In an older experiment of ours, with a trained encoder in Breakout, FC1 ended with a median of 53 of its 64 units alive and FC2 with 56, by the 0.01 rule. Our pre-set test for a concentration of dead units in FC1 required FC2 to keep at least 2.0 times as many units as FC1. It kept 1.06 times as many. Two cautions apply to this comparison. The silencing check failed in every variant of that older experiment. Their agents also use a value-based method, which learns the future reward of each action and then acts on it, and ours use PPO. We did not test value-based agents.

The experiment in four settings has one run per setting and encoder, with the exceptions listed in the table. We did not run the silencing check of Section 3.3 on its agents (Supplement S15, S4.6, S4.7 and S18.3).

\section{More live units did not always mean better learning}

This section presents the fourth finding. Dormant units are often treated as a fault to be fixed, for example by resetting them \citep{sokar2023dormant}. On that view one would expect an agent with more live units to learn better, since more of its network is in use. In our agents with a frozen encoder, that was not always the case. The finding is limited to those agents. Other studies have already reported that dormant units and performance do not always go together, as we describe below. What we add is an experiment in which we raise the number of live units on purpose and look at what happens to learning. The experiment has a weakness, which we describe below.

The first piece of evidence comes from the learning rate. If large steps are what make the units of FC1 die, then smaller steps for FC1 should leave more of them alive. We tested this by dividing the learning rate of FC1 alone by 49, which is the ratio of the fan-ins of the two dense layers (3,136 / 64), or by 7, which is the square root of 49. Every other layer kept the usual learning rate. More units did stay alive. In Breakout, by the 1 percent rule, the median number of live units was 14.5 in the comparison runs at the usual rate, 36 with the rate divided by 7, and 43 with the rate divided by 49. These gains are well beyond the largest difference we saw between repeated runs, which was 7 units. But the agents did not learn better. With the rate divided by 49, the Breakout agents ended with 36 to 46 live units. Even so, their average training scores over the run were lower than those of every run at the usual rate. In Pong, dividing the rate by 49 raised the number of live units from between 3 and 6 to between 34 and 41, and no agent reached a training score of 0 in 50 million agent steps. At the usual rate, five of the six agents did not reach it either.

This test has a weakness. A smaller learning rate may also slow down what FC1 can learn, which we did not test, so the lower scores do not show that the extra units were of no use. We also did not tune the reduced learning rate, and there is one run per setting and encoder (Supplement S12.2).

The second piece of evidence comes from centring in Breakout. Centring raised the number of live units from between 9 and 26 to between 30 and 37. These counts use the 1 percent rule, and the first range covers the 18 unaltered runs of the six encoders. We could not show that the centred agents scored any better. For each centred agent, the average training score over the run was within $-5.385$ to $+8.892$ points of the median of its encoder's unaltered runs. That is less than the largest difference we saw between repeated runs, which was 16.45 points. The centred agents did reach a training score of 10 sooner, by a factor of 3.300 to 4.895.

The image classifiers gave a third example. There, raising the offset reduced the number of live units while accuracy went up (Section 4.3).

In Pong, the number of live units and the speed of learning rose together. The agents with the lower offset had more live units and also learned much faster, although the speed of learning was not part of a rule we had set in advance (Section 4.2). Every centred agent reached a training score of 20 within 4.3 to 6.7 million agent steps. With the shift after the encoder, which lowers the offset without changing the screens, all six agents ended with 56 to 63 live units and reached a training score of 0 within 25 million agent steps. But both changed when we changed the offset, and these experiments cannot separate the harm done by having few live units from the harm done by the offset itself. The same caution applies to a comparison between frozen and trained encoders. There too the number of live units and the score are higher together (Section 6.2), but those networks differ in much more than their number of live units.

Two results suggest the opposite, that the number of live units does matter for performance, and we report them for completeness. Three Pong agents of the earlier study ended with 5, 7 and 11 live units, by the 0.01 rule, and scored +14, +18 and +19. Three runs are too few to set against the tests above. The second result comes from a side study in a different setting. We trained 420 networks with one hidden layer, 24 to 128 units wide, on 600 raw images of handwritten digits. The error on held-out images was more closely related to the logarithm of the number of live units than to the logarithm of the width of the layer (R-squared 0.69 against 0.59) (Supplement S4.3).

Other researchers have also found that dead or dormant units and performance do not always go together. Lyle et al. \citeyearpar{lyle2023plasticity} showed that the number of dead units does not reliably predict how much plasticity a network has lost (Section 2). He \citep{he2026plasticity}, Lee et al. \citeyearpar{lee2026rewardinflation} and Bu et al. \citeyearpar{bu2026rankcollapse} describe settings in which the number of dormant units and an agent's reward do not move together. Kooi et al. \citeyearpar{kooi2024hadamard} found that having fewer dormant units did not mean better play.

The reverse of this section's title also needs saying. Having few live units did not rule out playing well. The Pong agents of the earlier study played at expert level with between 3 and 8 live units of the 64, and its Breakout agents, with sticky actions off, reached 221 to 351 points with between 20 and 23 (Section 6.1). Those agents were trained for 100 to 200 million agent steps in Pong and for 400 to 853 million in Breakout. In our own Pong runs on unaltered screens, which lasted 50 million agent steps, the agents ended with between 3 and 6 live units, and none of them reached a training score of 18 (Section 6.2). So in our Pong runs a low number went with slow learning, although, as we said above, these runs cannot separate the effect of few live units from the effect of the offset. The two sets of counts also use different counting rules and different screens (Section 6.1), so we compare them only roughly. A network with few live units is therefore not, for that reason alone, a failed one. The runs of the earlier study suggest that whether it succeeds can depend on how long it is trained. We did not test this ourselves, because we did not train our own agents on unaltered screens for longer. We add one caution. Five other Pong agents of the earlier study ended with only 1 or 2 live units, and their scores stayed below zero. Those agents were trained in a different framework and with sticky actions fixed at 0.25, so we cannot say that the low number was the reason (Supplement S9).

This leaves the question of what the number of live units is useful for. In our agents a low number went with a large offset or a large step. Raising the number by making the step smaller did not improve learning. So we treat a low number of live units as a reason to check the offset and the step size, and not as something to be raised for its own sake (Supplement S14.2).

\section{Discussion}

Before turning to practice, we say in one place what the paper adds. Earlier work had shown that inputs far from zero leave ReLU units dead at the start of training in sparse autoencoders. We found that a higher offset also leaves fewer units alive at the end of training, in sparse autoencoders and in agents, image classifiers and a reservoir computer. In the sparse autoencoders, most of the units that ended dead at high offsets had been alive at the start, so they died during training. Earlier work had located dormant units in the layer after the encoder without testing why they are there. For the layer after a frozen encoder, we propose two quantities that the number of live units depends on, and both can be measured before training. The evidence is stronger for the offset than for the step size (Section 4.4). Earlier work had found frozen random encoders to learn poorly or slowly. Here, a random encoder left the offset of its input roughly unchanged, so a game with a bright, mostly unchanging screen gives the next layer a large offset. We gave a formula that predicts that offset. In our Pong runs, agents with a frozen encoder did not reach a training score of 18 in 50 million agent steps on unaltered screens. With the average screen subtracted, they reached it within 5 million. In Breakout the subtraction did not close the gap to a trained encoder. Last, in our agents with a frozen encoder, raising the number of live units did not always improve learning, so we do not treat that number as something to be raised for its own sake. A low number did not rule out success either. With much longer training, agents with between 3 and 8 live units reached expert play in Pong (Section 6.1). Each of these findings has limits, which Section 8.3 collects.

\subsection{What to do before training a layer on fixed features}

This subsection turns the findings into practical steps for a layer of ReLU units whose inputs stay fixed. The evidence behind the steps comes from the systems of Sections 4 to 7, and each step says how far it was tested.

\begin{enumerate}[noitemsep,topsep=2pt]
\item Measure the offset. Take a sample of the inputs to the layer and arrange them as a table with one row per example. Compute the length of the average row, and divide it by the square root of the sum of the variances of the columns (Section 3.2). One pass over the sample is enough. If the front end is a random encoder, the formula of Section 5.3 gives the median offset from the inputs and the layer sizes alone, before any weights are drawn. That formula gives the median over random encoders. It does not give the offset of one particular encoder.
\item If the offset is large, subtract the average input. The average is computed once, over the data. This is the remedy we tested most. In Breakout agents it raised the median number of live units from 16 to 27, by the 0.01 rule. In Pong agents it raised the number from between 3 and 6 to between 59 and 63, by the 1 percent rule (Section 4.2), and the agents reached a training score of 18 within 5 million agent steps, which none of the six unaltered agents did in 50 million. In sparse autoencoders on GPT-2, at a learning rate of 0.003, it lowered the fraction of dead units from 0.407 to 0.005 (Section 4.3). We cannot give a general threshold for what counts as a large offset. In our Breakout agents, the median number of live units fell steeply between offsets of 3.06 and 3.34, measured on the mixed set of screens, and not at one single offset (Section 4.2).
\item If a normalising layer is used, check which average it subtracts and where it is placed. Layer normalisation subtracts the average of each input separately and then rescales, which is a different operation from subtracting the average over the data. In image classifiers, layer normalisation applied to the encoder's outputs helped at only one of two bias values. Applied to the input sums of a layer, immediately before the ReLU, it kept nearly every unit alive (Section 4.3). We tested this in image classifiers only.
\item In our agents, a smaller learning rate for the layer with many inputs kept more of its units alive but did not help the agents learn (Section 7).
\item Apply the remedy from the start of training. Most of the units that are lost are lost early. In Breakout, most were gone by 2 million agent steps, counting the units that were already dead at the start (Supplement S13.2). This advice is an inference from that timing, because we did not test a remedy applied later in training.
\item When reporting the number of dead or dormant units, also report the counting rule, the inputs on which the units were counted, the point in training, and the offset of the layer's inputs (Supplement S16.1 and S18.2).
\end{enumerate}

\subsection{Where this applies, and where it is already handled}

Some training pipelines already subtract a mean or normalise. DrQ-v2 \citep{yarats2022drqv2} and DreamerV3 \citep{hafner2025dreamerv3} do so before the first nonlinearity that comes after the front end, and SimBa \citep{lee2025simba} normalises its observations (Section 2). If our explanation is right, such pipelines should not have the problem, but we did not test them. Other implementations only divide the pixel values by 255, with no centring and no normalising layer. The Atari PPO code in CleanRL \citep{huang2022cleanrl} and in Stable-Baselines3 \citep{raffin2021stable} is of this kind, and our own agents were trained with the second of these (Section 3.1). With such an implementation, the layer after a frozen encoder can receive inputs with a large offset, as it did in our agents. In our runs a trained encoder lowered its own offset early in training (Section 6.4). We did not test whether this is the reason such implementations work without centring when the encoder is trained.

Outside games, our own tests are those of Section 4.3: small image classifiers, a reservoir computer, and sparse autoencoders on GPT-2, whose fixed front end is a pretrained model and not a random one. We now list four other kinds of system in which the offset may matter. We tested none of them, and we mention them as places to look.

\begin{itemize}[noitemsep,topsep=2pt]
\item A layer of ReLU units trained on a frozen pretrained encoder. Two studies in robot learning place a batch-normalisation layer between the frozen features and the trained network, and they do not say why \citep{parisi2022unsurprising,nair2022r3m}. An offset in the frozen features is one possible reason.
\item A layer of ReLU units trained on the output of a reservoir or on random features \citep{jaeger2001echo,rahimi2007random,huang2006extreme}. One method that classifies time series with fixed random convolution kernels already takes its biases from quantiles of the convolution output on a training example \citep{dempster2021minirocket}. This places the threshold of each unit inside the range of its inputs.
\item Fixed representations that have a large component in common. Word vectors share a large mean vector \citep{mu2018allbutthetop}, and randomly initialised networks map their inputs to vectors that point in nearly the same direction \citep{liang2022mind}.
\item A randomly initialised encoder that is used as a control for a pretrained one \citep{berger2026ecg,isiacik2026timeseries}. If the output of such a control goes into a trained layer of ReLU units, and that output is not centred, the control may lose units for a reason that has nothing to do with what pretraining learned.
\end{itemize}

A frozen random encoder can also be used as a control in another kind of study. Some work draws conclusions from which units of a network are active. A frozen random encoder shows what the same layer does, on the same inputs, when nothing in the front end has been learned. Controls of this kind are already in use elsewhere. Some saliency methods change little when the upper layers of a trained model are randomised \citep{adebayo2018sanity}. Sparse autoencoders trained on randomly initialised transformers often score about as well on automated measures of interpretability as those trained on trained models \citep{heap2025interpretability}. We suggest the same kind of control for counts of active or dead units in layers that are trained on fixed features. We have not tried it on any model other than our own.

\subsection{Limits and open questions}

The first limit is the size of the evidence. All our networks are small. They are agents on three Atari games with dense layers of 64 units, small image classifiers, one reservoir computer, and one language model at one layer. Most conditions have three to six runs, one for each encoder, and runs that share an encoder are not independent of one another. We repeated the same encoder in the same setting only for one Breakout setting and, over the first 10 million agent steps, for two Pong settings. Those repeats showed that runs of one Breakout encoder can differ by up to 7 live units. In addition, the number of live units depends on the cut-off used for counting, on the set of screens, and on the point in training. For that reason we cannot say what share of the units are lost before training starts and what share during training.

Second, several of our pre-set tests failed or passed only in part. In the first Breakout experiment, the rank correlation between the offset and the number of live units fell short of the level we had set (Section 4.2). The tests for a sudden drop in the Breakout count failed, which is why we describe a ramp. No single rule for scaling the learning rate with the fan-in kept the number of live units constant, and in the sparse autoencoders the pre-set test of that scaling failed (Section 4.4). The silencing check failed in three experiments, and it was not run on the Pong agents of Section 4.2 or on the agents of Sections 6 and 7. For those agents we have not shown that the units counted as alive are the ones that matter for play (Section 3.3). Supplement S5 and S18 list every such test with its outcome.

Third, our changes to the offset also changed the overall size of the layer's inputs, and we made no correction for this. In the agents it follows from how the changes are made: a constant added to the encoder's biases makes its outputs larger, and a shift after the encoder changes the size of its outputs along with their offset. We measured the size only in an older Breakout experiment with a trainable encoder, where subtracting the average screen reduced the root-mean-square size of FC1's inputs at the start of training from a median of 0.2890 to 0.0875 over five seeds, and in the image classifiers, where it changed as well (Supplement S18.5). We have not checked it for the reservoir computer. These experiments therefore do not separate the effect of the offset from the effect of the size of the inputs. Larger inputs also let a training step move a unit's input sum further (Supplement S12.1), so part of what we attribute to the offset may act through the step size. The sparse autoencoders are the exception. There we raised the offset while keeping the overall size of the inputs fixed, and the fraction of dead units still rose (Section 4.3).

Fourth, some questions were not tested. We did not test whether the two quantities compound, whether the formula predicts the number of live units, or whether the fall in a trained encoder's offset is the reason its units stay alive. We did not test whether removing the rest of the offset in Breakout would close the gap to a trained encoder (Section 6.3). Finally, we do not know whether the difficulty of the task changes the number of live units in agents. Our one experiment designed to test this, with sticky actions on and off in Breakout, did not settle the question.

\section{Conclusion}

We asked what a fixed front end does to the layer that is trained on top of it. Earlier work had shown that inputs far from zero leave units dead at the start of training in sparse autoencoders, and had found dormant units in the layer after the encoder of reinforcement-learning agents without testing the cause. In the systems we tested, the output of the front end can lie far from zero compared with how much it varies, and we call this ratio the offset. The number of units still alive at the end of training depended on the offset. In our agents it also appeared to depend on the size of the training steps, on weaker evidence.

Four findings support this. Raising the offset left fewer units alive in agents, in image classifiers and in a reservoir computer, and more units dead in sparse autoencoders on GPT-2. In Pong, subtracting the average screen raised the number of live units from between 3 and 6 of the 64 to between 59 and 63, by our main counting rule. A random encoder leaves the offset of its input roughly unchanged, and a formula based on the inputs and the layer sizes predicted its median offset to within 25 percent in all 44 conditions. With the average screen subtracted, a frozen encoder learned Pong about as fast as a trained encoder did on unaltered screens, within the difference between repeated runs. It stayed far behind in Breakout, and a trained encoder lowered its own offset early in training. And in agents with a frozen encoder, having more live units did not always mean learning better.

The title asks what the number of live units measures. In our experiments it depended on the offset of the layer's inputs and, in our agents and on weaker evidence, on the size of the training steps. The task also changed it in image classifiers, but downwards with one kind of output layer and upwards with another. So it was not a consistent measure of the task, as our earlier study had taken it to be. It also did not always go together with how well the network learned. A network with few live units could still learn its game well. In the earlier study, Pong agents with between 3 and 8 live units of the 64 reached expert play, after much longer training.

Within the settings we tested, we suggest three things. Measure the offset of a layer's inputs before training. If it is large, try subtracting the mean of the inputs first. And when reporting how many units are dead or dormant, report the offset as well, because two networks can have different numbers of dead units without differing in how well they learned. The evidence comes from small networks, three Atari games and, for the most part, one run per setting and encoder, and the tests outside games are small. In the agents and the image classifiers, our changes to the offset also changed the size of the layer's inputs.

\phantomsection\addcontentsline{toc}{section}{Use of AI tools}
\section*{Use of AI tools}

This work was done with AI assistants, mainly Anthropic's Claude models used through Claude Code. They wrote and ran most of the code, drafted the text and made the figures, under the author's direction and review. The author is responsible for everything in the paper.

\bibliography{refs_2026-11-21}
\bibliographystyle{tmlr}

\end{document}